\documentclass[10pt,twocolumn,letterpaper]{article}

\usepackage{iccv}
\usepackage{times}
\usepackage{epsfig}
\usepackage{graphicx}
\usepackage{amsmath}
\usepackage{amssymb}
\usepackage{xcolor}
\usepackage{bbm}
\usepackage{bm}
\usepackage[numbers,sort&compress]{natbib}  % Compress citations and order them
\usepackage{multirow}  % For \multirow in tables
\usepackage{subcaption}
\usepackage{dutchcal}
\usepackage{arydshln}  % For \hdashline
\usepackage{booktabs}  % For \toprule and \bottomrule
\usepackage{xspace}
\usepackage{enumitem}
\usepackage[accsupp]{axessibility} % For accessability
\usepackage[pagebackref=true,breaklinks=true,letterpaper=true,colorlinks,bookmarks=false]{hyperref}

\iccvfinalcopy % *** Uncomment this line for the final submission

\makeatletter
\def\ps@myheadings{%
    \let\@oddfoot\@empty\let\@evenfoot\@empty
    \def\@evenhead{\thepage\hfil\slshape\leftmark}%
    \def\@oddhead{{\slshape\rightmark}\hfil\thepage}%
    \let\@mkboth\@gobbletwo
    \let\sectionmark\@gobble
    \let\subsectionmark\@gobble
    }
  \if@titlepage
  \renewcommand\maketitle{\begin{titlepage}%
  \let\footnotesize\small
  \let\footnoterule\relax
  \let \footnote \thanks
  \null\vfil
  \vskip 60\p@
  \begin{center}%
    {\LARGE \@title \par}%
    \vskip 3em%
    {\large
     \lineskip .75em%
      \begin{tabular}[t]{c}%
        \@author
      \end{tabular}\par}%
      \vskip 1.5em%
    {\large \@date \par}%       % Set date in \large size.
  \end{center}\par
  \@thanks
  \vfil\null
  \end{titlepage}%
  \setcounter{footnote}{0}%
}
\else
\renewcommand\maketitle{\par
  \begingroup
    \renewcommand\thefootnote{\@fnsymbol\c@footnote}%
    \def\@makefnmark{\rlap{\@textsuperscript{\normalfont\@thefnmark}}}%
    \long\def\@makefntext##1{\parindent 1em\noindent
            \hb@xt@1.8em{%
                \hss\@textsuperscript{\normalfont\@thefnmark}}##1}%
    \if@twocolumn
      \ifnum \col@number=\@ne
        \@maketitle
      \else
        \twocolumn[\@maketitle]%
      \fi
    \else
      \newpage
      \global\@topnum\z@   % Prevents figures from going at top of page.
      \@maketitle
    \fi
    \thispagestyle{plain}\@thanks
  \endgroup
  \setcounter{footnote}{0}%
}
\makeatother
\ificcvfinal\fi

\begin{document}
\newcommand{\AS}[1]{{\color{red}[AS: #1]}}
\newcommand{\AD}[1]{{\color{cyan}[AD: #1]}}
\newcommand{\OH}[1]{{\color{blue}[OH: #1]}}
\newcommand{\TL}[1]{{\color{green}[TL: #1]}}

\newcommand{\todo}[1]{{\color{magenta}\textbf{TODO}: #1}}
\newcommand{\XW}[1]{{\color{orange}[XW: #1]}}
\newcommand{\oh}[1]{{\OH{#1]}}}
\definecolor{xucongcolor}{rgb}{0.73725, 0.6588, 0.0705}
\newcommand{\xucong}[1]{\textsf{\textcolor{xucongcolor}{\textbf{Xucong:} \textit{#1}}}}

\let\biconditional\leftrightarrow
\newcommand{\head}[1]{\noindent\textbf{#1}}

\newcommand{\figref}[1]{Fig.~\ref{#1}}
\newcommand{\tabref}[1]{Tab.~\ref{#1}}
\newcommand{\V}[1]{\mathbf{#1}}
\newcommand{\R}[0]{\rm I\!R}
\newcommand{\E}[0]{\rm I\!E}
\newcommand{\loss}[0]{\mathcal{L}}

\newcommand{\Rone}{\textbf{\textcolor{NavyBlue}{R1}}}
\newcommand{\Rtwo}{\textbf{\textcolor{ForestGreen}{R2}}}
\newcommand{\Rthree}{\textbf{\color{RedOrange}{R3}}}
\newcommand{\A}{\textbf{A:}}

\newcommand{\methodname}{PeCLR\xspace}

\newcommand{\myparagraph}[1]{\noindent\textbf{#1.}}

\newcommand{\newtext}[1]{{\color{magenta}\textbf{NEW}: #1}}

\newcommand\blfootnote[1]{%
  \begingroup
  \renewcommand\thefootnote{}\footnote{#1}%
  \addtocounter{footnote}{-1}%
  \endgroup
}

%%%%%%%%% TITLE
% \title{Contrastive Learning improves Hand Pose Estimation}
% Self-supervised 3D Hand Pose Estimation from monocular RGB through Contrastive Learning
% Equivariant contrastive learning for Hand pose estimation.
% CLHand: Contrastive Learning for Hand Pose Estimation.
\title{PeCLR: Self-Supervised 3D Hand Pose Estimation from monocular RGB via Equivariant Contrastive Learning}

\author{Adrian Spurr* \quad Aneesh Dahiya* \quad Xi Wang \quad Xucong Zhang \quad Otmar Hilliges\vspace{0.1cm} \\
Department of Computer Science, ETH Zurich, Switzerland\\
}
\date{}
\maketitle
% Remove page # from the first page of camera-ready.

\ificcvfinal\thispagestyle{empty}\fi

%%%%%%%%% ABSTRACT
\begin{abstract}
Encouraged by the success of contrastive learning on image classification tasks, we propose a new self-supervised method for the structured regression task of 3D hand pose estimation.
Contrastive learning makes use of unlabeled data for the purpose of representation learning via a loss formulation that encourages the learned feature representations to be invariant under any image transformation.
For 3D hand pose estimation, it too is desirable to have invariance to appearance transformation such as color jitter. 
However, the task requires equivariance under affine transformations, such as rotation and translation.
To address this issue, we propose an equivariant contrastive objective and demonstrate its effectiveness in the context of 3D hand pose estimation.
We experimentally investigate the impact of invariant and equivariant contrastive objectives and show that learning equivariant features leads to better representations for the task of 3D hand pose estimation. 
Furthermore, we show that standard ResNets with sufficient depth, trained on additional unlabeled data, attain improvements of up to $14.5\%$ in PA-EPE on FreiHAND and thus achieves state-of-the-art performance without any task specific, specialized architectures. Code and models are available at \href{https://ait.ethz.ch/projects/2021/PeCLR/}{https://ait.ethz.ch/projects/2021/PeCLR/}
\blfootnote{*Denotes equal contribution}
\end{abstract}

\begin{figure}[h]
  \centering
  \includegraphics[width=\linewidth]{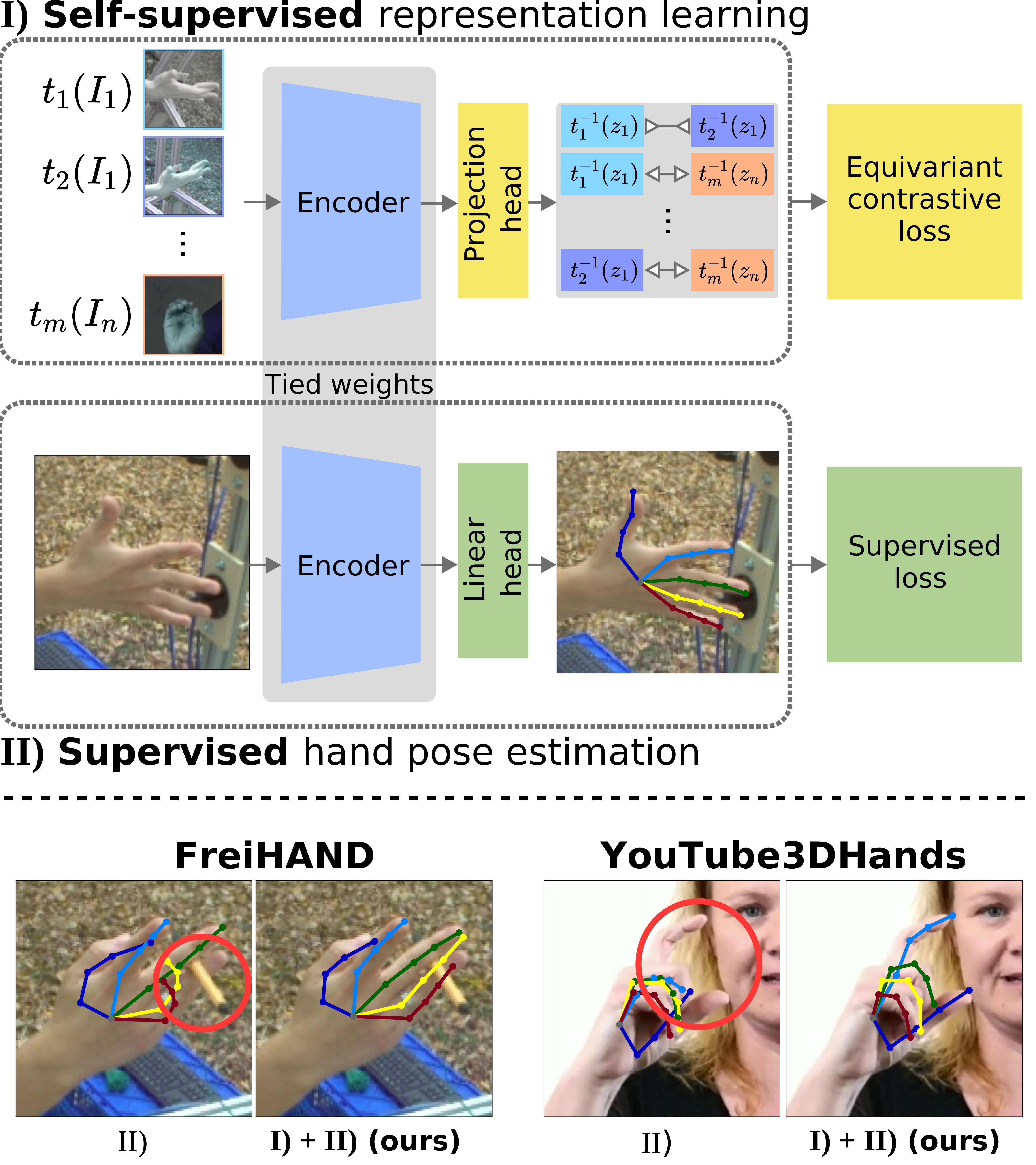}
   \caption{We propose a two-stage framework for 3D hand pose estimation. I) An encoder is trained in a self-supervised manner on a large set of unlabeled data using a novel equivariant contrastive objective. II) The pre-trained encoder is fine-tuned with little labeled data. The resulting network is more accurate across datasets.}
\label{fig:teaser}
\end{figure}

%%%%%%%%% BODY TEXT
\section{Introduction}
\label{sec:intro}
Estimating the 3D pose of human hands from monocular images alone has many important applications in robotics, Human-Computer Interaction and AR/VR. As such the problem has received significant attention in computer vision literature~\cite{zb2017hand, spurr2018cross, hasson2019learning, tekin2019ho, doosti2020hope, spurr2020weakly, moon2020interhand2, hasson2020leveraging}. 
However, estimating the location of 3D hand joints within an RGB image is a challenging structured regression problem with difficulties that arise from a large diversity in backgrounds, lighting conditions, hand appearances, as well as self-occlusion caused by the high degrees of freedom of the human hand.

Annotated datasets that cover a larger diversity of environments and settings are one possibility to alleviate this issue. However, acquiring 3D labeled data is laborious, cost intensive and typically requires multi-view imagery or some form of user instrumentation. Data collected under such circumstances is often difficult to transfer well to in-the-wild imagery \cite{zimmermann2019freihand,youtubehand}. Therefore, much interest is given to approaches that can leverage auxiliary data, which has either no or only 2D joint annotations.
For example, such data can be used to outperform many supervised approaches via making use of weak-supervision \cite{cai2018weakly, boukhayma20193d}, the integration of kinematic priors \cite{spurr2020weakly}, or by exploiting temporal information \cite{hasson2020leveraging}.
Off-the-shelf joint detectors \cite{openpose} have been leveraged to automatically generated 2D annotations in large quantities \cite{youtubehand}. However, the accuracy of models trained on these labels, or on 3D annotations derived from them, are inherently bounded by the label noise. Therefore, the question of how to efficiently leverage unlabeled data for hand pose estimator training remains unanswered.

Recently, self-supervised approaches such as contrastive learning have shown that they can reach parity with supervised approaches on image classification tasks \cite{SimCLR, moco}. These methods leverage unlabeled data to learn powerful feature representations. To do so, positive and negative pairs of images are projected into a latent space via a neural network. 
The contrastive objective encourages the latent space samples of the positive pairs to lie close to each other and pushes negative pairs apart. 
The resulting pre-trained network can then be applied to downstream tasks. 
Positive pairs are created by sampling an image and applying two sets of distinct augmentations on it, whereas negative pairs correspond to separate but similarly augmented images. 
These augmentations include appearance transformations, such as color drop, and geometric transformations, such as rotation. 
The contrastive objective induces invariance under all of these transformations. 
However, 3D regressions tasks, such as hand pose estimation, inherently require \emph{equivariance} under \emph{geometric} transformations. Hence, representations learnt from a standard contrastive objective may not effectively transfer to pose estimation.

To the best of our knowledge, for the first time, we investigate self-supervised representation learning techniques for 3D hand pose estimation in this paper. 
We derive a method named \textit{Pose Equivariant Contrastive Learning (\methodname)}. 
One of our core contributions is a novel formulation of a contrastive learning objective that induces equivariance to geometric transformations and we show that this allows to effectively leverage the large diversity of existing hand images without \emph{any} joint labels.
These images are used to pre-train a network, which can then be transferred to the final hand pose estimation task via supervised fine-tuning. 
This provides a promising direction for hand pose estimation and enables an easy transfer of images collected in-the-wild or calibration to a specific domain by fine-tuning a pre-trained network with fewer labels.

\figref{fig:teaser} provides an overview of our method. First, we perform self-supervised representation learning. Given an RGB image of the hand, we apply appearance and geometric transformations to generate positive and negative pairs of derivative images. These are used to train an encoder via our proposed equivariant contrastive loss. By undoing the geometric transformation in latent space, we promote equivariance. However, invertion of these transformations is not straightforward. This is because transformations on images should lead to proportional changes in the latent space. Therefore special care needs to be taken due to different magnitudes between latent space and pixel space under learned projection. We propose a latent sample normalization technique that compensates for this difference and we show that the resulting model yields improved pose estimation accuracy (cf. \figref{fig:teaser}, bottom) compared to both supervised and standard contrastive learning.

In the second stage, the pre-trained encoder is fine-tuned on the task of 3D hand pose estimation using labeled data. The resulting model is evaluated thoroughly in a variety of settings. We demonstrate increased label efficiency for semi-supervision and show that using more unlabeled data is beneficial for the final performance, yielding improvements of up to $43\%$ in 3D EPE in the lowest labeled setting (cf. \figref{fig:semi_sup}). Next, we show that this improvement also transfers to the fully supervised case, where using a standard ResNet with sufficient depth in combination with unlabeled data and our proposed pre-training scheme outperforms specialized state-of-the-art architectures (cf. \tabref{tab:SOTAcompare}). Finally, we demonstrate that self-supervised pre-training leads to an improvement of $5.6\%$ 3D PA-EPE in cross-data evaluation, indicating that pre-training is beneficial for cross-domain generalization (cf. \tabref{tab:cross_dataset_analysis}).

In summary, our contributions are as follows:
\begin{enumerate}[noitemsep]
    \item To the best of our knowledge, we perform the first investigation of contrastive learning to efficiently leverage unlabeled data for 3D hand pose estimation.
    \item We propose a contrastive learning objective that encourages \emph{invariance} to appearance transformations and \emph{equivariance} to geometric transformations. 
    \item We conduct controlled experiments to empirically derive the best performing augmentations.
    \item We show that the proposed method achieves better label efficiency in semi-supervised settings and that adding more unlabeled data is beneficial.
    \item We empirically show that our proposed method outperforms current, more specialized state-of-the-art methods using standard ResNet models.
\end{enumerate}

\noindent Code and models are available for research purposes:  \href{https://ait.ethz.ch/projects/2021/PeCLR/}{https://ait.ethz.ch/projects/2021/PeCLR/}.

\section{Related work}
\noindent\textbf{Hand pose estimation.}
Hand pose estimation usually follows one of three paradigms. Some work predicts 3D joint skeletons directly \cite{zb2017hand, spurr2018cross, mueller2018ganerated, iqbal2018hand, cai2018weakly, tekin2019ho, yang2019disentangling, doosti2020hope, spurr2020weakly, moon2020interhand2}, make use of MANO \cite{romero2017mano}, where the parameters of a parametric hand model are regressed \cite{baek2019pushing, boukhayma20193d, hasson2019learning, baek2020weakly, hasson2020leveraging, zhang2019end}, or predicts the full mesh model of the hand directly \cite{ge20193d, kulon2020weakly, moon2020deephandmesh}. A staged approach is introduced in \cite{zb2017hand}, where the 2D keypoints are regressed directly and then lifted to 3D. Spurr \etal \cite{spurr2018cross} introduces a cross-modal latent space which facilitates better learning. Mueller \etal \cite{mueller2018ganerated} makes use of a synthetically created dataset and reduces the synthetic/real discrepancy via a GAN. Cai \etal \cite{cai2018weakly} makes use of supplementary depth supervision to augment the training set. Proposing a more efficient hand representation, a 2.5D representation is introduced in \cite{iqbal2018hand}. Action recognition as well as hand/object pose estimation is performed in \cite{tekin2019ho}. \cite{yang2019disentangling} introduces a disentangled latent space, for the purpose of better image synthesis. A graph-based neural network is used to jointly refine the hand/object pose in \cite{doosti2020hope}.  
Biomechanical constraints are introduced to refine the pose predictions on 2D supervised data \cite{spurr2020weakly}. Moon \etal \cite{moon2020interhand2} predict the pose of both hands and takes their interaction into account.

Templated-based methods such as MANO induce a prior of hand poses, as well as providing a mesh surface. Some methods \cite{baek2019pushing, boukhayma20193d, zhang2019end} estimate the MANO parameters directly from RGB, sometimes making use of weak supervision such as hand masks \cite{baek2019pushing, zhang2019end} or in-the-wild 2D annotations \cite{boukhayma20193d, zhang2019end}. A unified approach is introduced to jointly predict MANO as well as the object mesh \cite{hasson2019learning}. Hasson \etal \cite{hasson2020leveraging} builds upon the mentioned framework, by learning from partially labeled sequences via a photometric loss. An alternative to MANO is proposed in \cite{moon2020deephandmesh} by predicting pose and subject dependant correctives to a base hand model. 
Some methods regress the mesh of a hand directly. However, mesh annotations are difficult to acquire. Ge \etal \cite{ge20193d} tackles this by introducing a fully mesh-annotated synthetic dataset and performs noisy supervision for real data. With the help of spiral convolutions, a hand mesh is predicted in \cite{kulon2020weakly}, supervised using MANO. 

Clearly, much work has been dedicated to custom, sometimes highly specialized architectures for hand-pose estimation. In contrast, we explore a purely data-driven approach, utilizing unlabeled data, and an equivariance inducing contrastive formulation to achieve state-of-the-art performance with a standard CNN.

\noindent\textbf{Self-supervised learning.}
Self-supervised learning aims to learn representation of data without any annotations. Literature defines the pre-text task as the specific strategy to learn the representation in a self-supervised manner. Such tasks include predicting the position of a second patch relative to the first~\cite{doersch2015unsupervised}, colorizing a grayscale image~\cite{zhang2016colorful}, solving a jigsaw puzzle~\cite{noroozi2016unsupervised}, estimating the motion flow of pixels in a scene~\cite{walker2016uncertain}, predicting positive future samples in audio signals~\cite{oord2018representation}, or completing the next sentence based on relations between two sentences~\cite{devlin2019bert}. However, it is not clear which pretext task would be optimal given a specific downstream task in terms of performance and generalizability. 

Contrastive learning is a powerful paradigm for self-supervised, task-independent learning. At the core of contrastive learning lies a concept emerging from distance metric learning, where a pair of data is encouraged to be close in latent space if they are connected in a meaningful way, while unrelated data are pushed apart. One of the appeals of contrastive learning lie in the numerous amounts of data that is available for training. General representations are learned through this paradigm and have been successfully used in many downstream tasks such as image and video classification~\cite{tian2019contrastive, SimCLR, moco}, object detection~\cite{wu2018unsupervised, henaff2020dataefficient}, and speech classification~\cite{oord2018representation}. However, contrastive learning has not been investigated for the task of hand pose estimation.

Contrastive learning has been explore in works such as Contrastive Predictive Coding (CPC) \cite{oord2018representation, henaff2020dataefficient}, 
Contrastive Multiview Coding (CMC) \cite{tian2019contrastive}, and SimCLR \cite{SimCLR, chen2020big}.
CPC learns to extract representations by predicting future representations in latent space. Autoregressive models are used to enable predictions of many steps in the future. While CPC learns from the two views of the past and future, CMC extends this idea to multi-view learning. It aims to learn view-invariant representations by maximizing mutual information among different views of the same content. 
The most relevant framework for contrastive learning is a simple yet effective approach \cite{SimCLR}. It largely benefits from data augmentation and its learnt representation achieves performance that is on par with supervised models on the image classification task. However, the learned transformation-invariant features are not suited for structured regression tasks such as hand pose estimation as these require an equivariant representation with respect to geometric transformations. In this work, we extend SimCLR by differentiating between appearance and geometric transformations, and propose a model that can successfully learn representations dedicated for both transformations.

\section{Method}
\begin{figure*}
     \centering
     \begin{subfigure}[b]{0.5\textwidth}
        \centering
        \includegraphics[width=1.0\columnwidth]{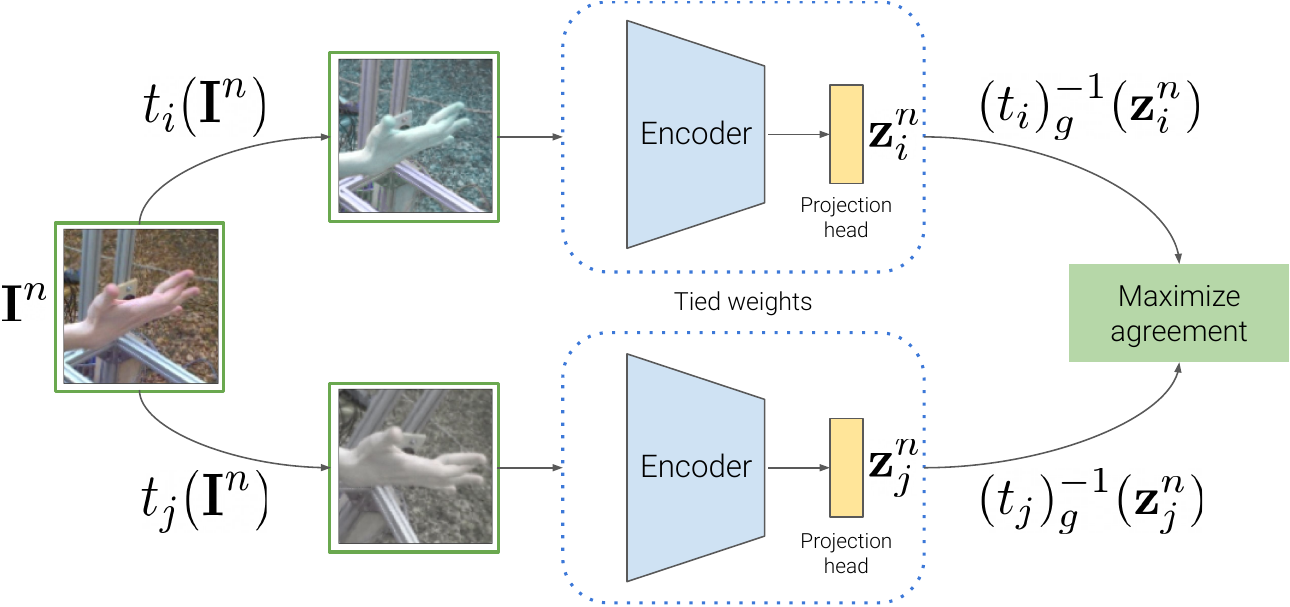}
        \label{fig:model_maximize}
     \end{subfigure}%
     \hfill
     \begin{subfigure}[b]{0.5\textwidth}
         \centering
        \includegraphics[width=1.0\columnwidth]{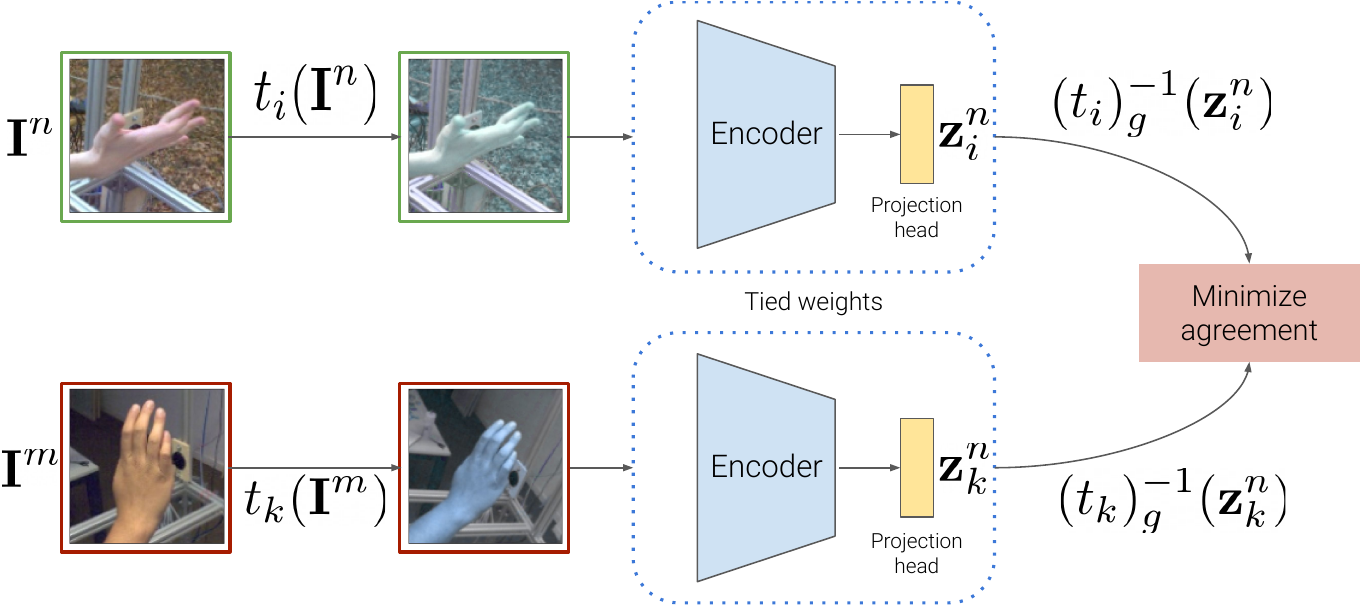}
         \label{fig:model_minimize}
     \end{subfigure}
   \caption{%
   \textbf{Method overview}. An augmentation $t =t_g \circ t_a$ is applied to input image $\bm{I}^n$. Here $t_g$ and $t_a$ denote the geometric and appearance components of the augmentation $t \in T$, respectively. The model then generates the projections $\bm{z}^n$ for each augmented input. Geometric augmentations are \emph{reversed} in \emph{projection space} before optimizing the contrastive objective. The agreement between projections from the same input image is maximized (left) and agreements amongst projections from different input images are minimized (right).}
\label{fig:model}
\end{figure*}

We start by reviewing SimCLR \cite{SimCLR}. We then introduce the overall framework of pre-training and finetuning. Next, we identify an issue with SimCLRs contrastive formulation when applied to hand pose estimation, motivating our proposed equivariant contrastive objective. %
Lastly, we present our hand pose estimation model and the method used for 3D keypoint estimation during supervised training.

\noindent\textbf{Notation.} In the following, we denote the set of all transformations used as $T$. It contains \emph{appearance} transformations $t^a$ (e.g color jitter), \emph{geometric} transformations $t^g$ (e.g. scale, rotation and translation) as well as compositions of them. For a given transformation $t_i \in T$, $t_i^a, t_i^g$ correspond to the appearance or geometric component of the transformation $t_i$. \figref{fig:augmentation} shows all transformation used in this study.

\subsection{SimCLR}
The idea of the SimCLR \cite{SimCLR} framework is to maximize the agreement in latent space between the representations of samples that are similar, while repelling dissimilar pairs. The positive pairs are artificially generated by applying various augmentations on an image. Given a set of samples $\{\bm{I}^n\}_{n=1}^N$, we consider two augmented views $\{\bm{I}_i^n, \bm{I}_j^n\}$, where $\bm{I}_i^n=t_i(\bm{I}^n)$, $\bm{I}_j^n=t_j(\bm{I}^n)$, $t_i, t_j \in T$.

The framework consists of an encoder $E$ and a projection head $g(\cdot)$. The overall model ${f} = g \circ E$ maps an image $\bm{I}$ to a latent space sample $\bm{z} \in \mathbb{R}^{k}$, i.e. $\bm{z}^n_i= f(\bm{I}_i^n)$. It is trained using a contrastive loss function that maximizes the agreement between all positive pairs of projections $\{\bm{z}_i^n, \bm{z}_j^n\}_{i \neq j}$, which are extracted from two augmented views of the same image $\bm{I}^n$. Simultaneously, it also minimizes the agreement amongst negative pairs of projections $\{\bm{z}_i^n, \bm{z}_k^m\}$, where $\bm{z}_k^m$ are extracted from different images.

In each iteration, SimCLR samples both positive and negative pairs. For a given batch of $N$ images, two augmentations are applied on each sample, resulting in $2N$ augmented images. Hence, for every augmented image $\bm{I}_i^n$, there is one positive sample $\bm{I}_j^n$, and $2(N-1)$ negative samples $\{\bm{I}_k^m\}_{m \neq n}$. The model is trained to project positive samples close to each other, whereas keeping negative samples far apart. This is achieved via the following loss function, termed as \emph{NT-Xent} in \cite{SimCLR}:
\begin{equation}
\label{eq:contrastive}
    \loss_{i,j} = - \log \frac{\exp{(\text{sim}(\bm{z}_i, \bm{z}_j)/\tau)}}{\sum_{k=1}^{2N} \mathbbm{1}_{[k\neq i]} \exp{(\text{sim}(\bm{z}_i,\bm{z}_k)/\tau)}}
\end{equation}
Here $\tau$ is a temperature parameter, $\text{sim}(\bm{u}, \bm{v}) = \bm{u}^T\bm{v} / ||\bm{u}||||\bm{v}||$ is the cosine similarity between $\bm{z}_i^n$, $\bm{z}_j^n$ and $\mathbbm{1}_{[k\neq i]}$ is the indicator function.

\subsection{Equivariant contrastive representations}
\label{sec:peclr}
Inspecting Eq. \ref{eq:contrastive}, we observe that the objective function promotes invariance under all transformations. Given a sample $\bm{I}^n_j = t_j(\bm{I}^n)$ and its positive sample $\bm{I}^n_i = t_i(\bm{I}^n) = t_i (t_j^{-1}(\bm{I}^n_j)) = \tilde{t}_i(\bm{I}^n_j)$, the numerator in Eq. \ref{eq:contrastive} is minimized if $f(\bm{I}^n_j) = \bm{z}^n_j = \bm{z}^n_i = f(\tilde{t}_i(\bm{I}^n_j))$. Hence, a model that satisfies Eq. \ref{eq:contrastive} needs to be invariant to all transformations in ${T}$. However, hand pose estimation requires equivariance with respect to geometric transformations as these change the displayed pose. Hence, we require:
\begin{equation}
\label{eq:equivariance}
    t^g_i f(\bm{I}^n_j) = f(t^g_i (\bm{I}^n_j)).
\end{equation}

\noindent\textbf{Inverting transformations in latent space.}
To fulfill Eq. \ref{eq:equivariance}, we first note that it is equivalent to $f(\bm{I}^n_j) = (t^g_i)^{-1}  f(t^g_i (\bm{I}^n_j)) \biconditional \bm{z}^n_j = (t^g_i)^{-1} \bm{z}^n_i$.
This leads us to the following equivariant modification of \emph{NT-Xent}:
\begin{equation}
    \loss_{i,j} = - \log \frac{\exp{(\text{sim}((\bm{\tilde{z}}_i, \bm{\tilde{z}}_j)/\tau)}}{\sum_{k=1}^{2N} \mathbbm{1}_{[k\neq i]} \exp{(\text{sim}(\bm{\tilde{z}}_i,\bm{\tilde{z}}_k)/\tau)}},
    \label{eq:contrastive_equiv}
\end{equation}
where $\bm{\tilde{z}}_i = (t^g_i)^{-1}\bm{z}_i$ and $\bm{z}_i \in \mathbb{R}^{m \times 2}$. In order to minimize the numerator in Eq. \ref{eq:contrastive_equiv} it must hold that $\bm{\tilde{z}}_i = \bm{\tilde{z}}_j$, which leads to the desired property of Eq. \ref{eq:equivariance}. Further details can be found in the supplementary. As $t^g_i$ is an affine transformation, its inverse can be easily computed. However, whereas scaling and rotation are transformations that are performed relative to the image size, translation is performed in terms of an absolute quantity. In other words, if we translate an image $\bm{I}^n$ by $x$ pixels, we need to translate its latent space projection $\bm{z}^n$ by a proportional quantity. 
Therefore, we translate $\bm{z}^n$ by a quantity proportional to its magnitude. To achieve this, we obtain the translation proportional to the image size and scale it up by a factor proportional to the range spanned by the projections in latent space.
To this end, we normalize the translation vector $\bm{\hat{v}}$ before applying its inverse to a latent space sample $\bm{z}_i$ to undo the transformation. The normalized vector $\bm{\hat{v}}$ is computed as follows:
\begin{equation}
    \bm{\hat{v}} = \frac{\bm{v}}{L} L_z
\end{equation}
Where $L_z = \text{max}(\bm{z}_i) - \text{min}(\bm{z}_i)$ and $L$ is the image length. The intuition behind $L_z$ is that it corresponds to the magnitude of latent space values. Hence, the resulting translation vector is proportional in magnitude.
Lastly, we note here that due to the cosine similarity used in Eq. \ref{eq:contrastive_equiv}, the effect of scaling is effectively removed (\ie $\text{sim}(a\bm{z}_i,b\bm{z}_j) = \text{sim}(\bm{z}_i,\bm{z}_j)$, for $a,b\in\mathbb{R}$). The complete equivariant contrastive learning framework is visualized in Fig.~\ref{fig:model}.

\noindent\textbf{From pre-training to fine-tuning.}
After having performed pre-training using our proposed loss function, we fine-tune the encoder supervised on the task of hand pose estimation. To this end, following \cite{SimCLR} we remove the projection layer $g$ from the model and replace it with a linear layer. The entire model is then trained end-to-end using the losses as described next, in Sec. \ref{sec:method_hpe}.

\subsection{3D Hand Pose Estimator}
\label{sec:method_hpe}
Our hand pose estimation model makes use of the 2.5D representation \cite{iqbal2018hand}. Given an image, the network predicts the 2D keypoints $\bm{J}^{2D} \in \R^{21\times2}$ and the root-relative depth $\bm{d}^r \in \R^{21}$ of the hand. As such, our hand pose model is trained with the following supervised loss functions:
\begin{equation}
    \begin{split}
    \label{eq:loss_25d}
    \loss_{\bm{J}^{2D}} &= |\bm{\hat{J}}^{2D} - \bm{J}^{2D}| \\
    \loss_{\bm{d}^r} &= |\mathbf{\hat{d}}^r - \bm{d}^r| 
    \end{split}
\end{equation}
Given the predicted values of $\mathbf{J}^{2D}$ and $\bm{d}^r$, the depth value of the root keypoint $d^{root}$ can be acquired as detailed in \cite{iqbal2018hand}. As a final step, we refine the acquired root depth to increase accuracy and stability as described \cite{spurr2020weakly}, which yields $d^{\mathrm{root}}_{\mathrm{ref}}$. The resulting 3D pose is acquired as follows:
\begin{equation}
    \begin{split}
    \label{eq:25d3d}
    \V{J}^{3D} = \V{K}^{-1} \V{J}^{2D} (\V{\bm{d}}^r + d^{\mathrm{root}}_{\mathrm{ref}}),
    \end{split}
\end{equation}
where $\V{K}$ is the camera intrinsic matrix.

\section{Experiments}
\label{sec:experiments}
Sec.~\ref{subsec:Importance of augmetations} investigates the impact of different data augmentation operations and evaluate their effectiveness in the hand pose estimation task. Next, with the self-supervised learnt representation, we demonstrate in Sec.~\ref{subsec:exp_semi_supervised} how our model efficiently makes use of labeled data in semi-supervised settings.
In Sec.~\ref{subsec:sota_comparison} we compare our method with related works in hand pose estimation and demonstrate that \methodname can reach state-of-the-art performance on FH. Finally, in Sec.~\ref{subsec:exp_cross_dataset} we perform a cross-dataset evaluation to show the advantages of the proposed representation learning across domain distributions.
\begin{figure*}
     \centering
    \includegraphics[width=2\columnwidth]{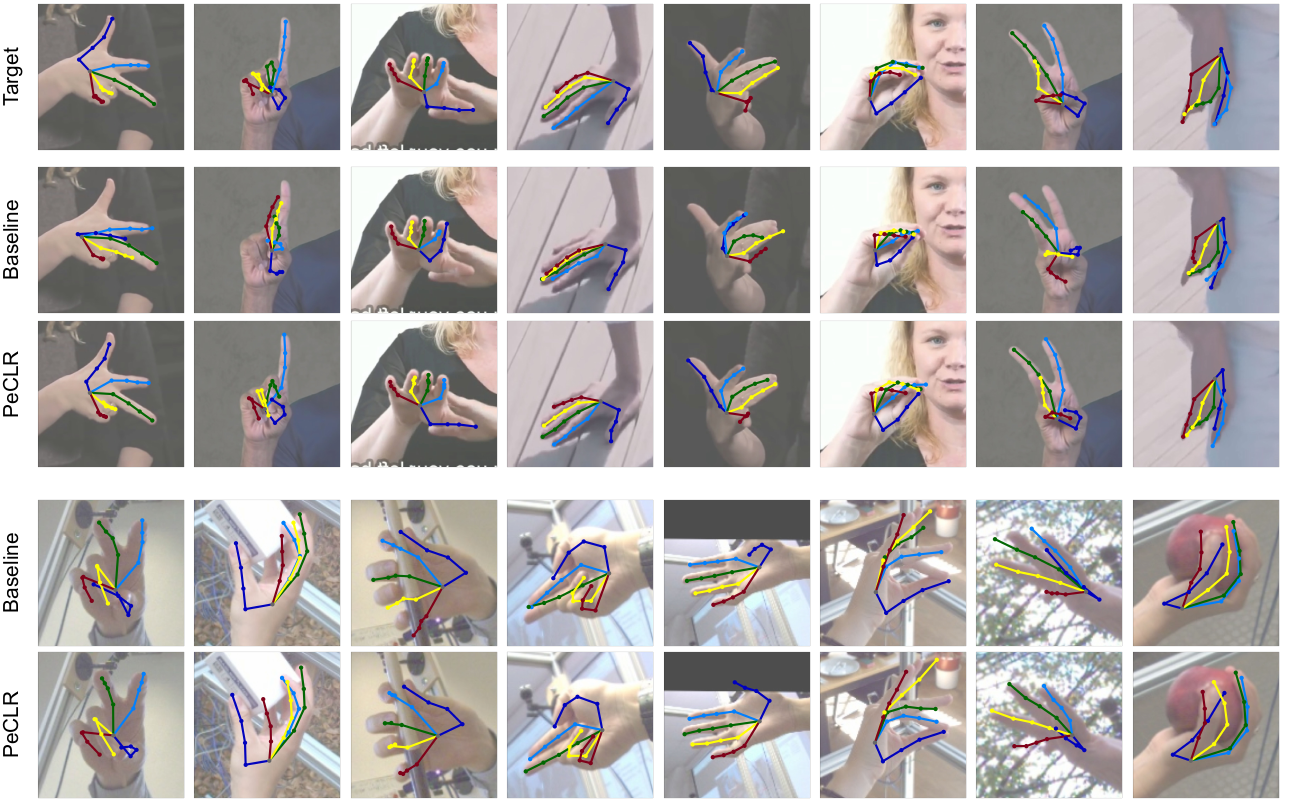}
    \caption{Predictions are shown on the test sets of YT3D (top) and FH (bottom) using either RN152 (Baseline) or RN152 + PeCLR. Note that the ground truth of the test set is not publicly available for FH, thus we only visualize the predictions.}
\label{fig:qual_results}
\end{figure*}

\subsection{Implementation}
For pre-training, we use ResNet (RN) \cite{he2016deep} as encoder, which takes monocular RGB images of size $128 \times 128$ as input. We use LARS \cite{LARS} with ADAM \cite{kingma2014adam} with batches of size 2048 and learning rate of $4.5e$-$3$ in the representation learning stage. During fine-tuning, we use RGB images of size $128 \times 128$ (Sec. \ref{subsec:Importance of augmetations}, \ref{subsec:exp_semi_supervised}) or $224 \times 224$ (Sec. \ref{subsec:sota_comparison}, \ref{subsec:exp_cross_dataset}). As optimizer we use ADAM with a learning rate of $5e$-$4$ in the supervised fine-tuning stage. Further training details can be found in the supplementary.

\subsection{Evaluation Metrics}
We report the End-point-error (EPE) and the Area-Under-Curve (AUC). EPE denotes the average euclidean distance between the ground-truth and predicted keypoints. AUC denotes the area under the Percentage-of-correct-Keypoints (PCK) curve for threshold values between 0 and 5 cm in 100 equally spaced increments. Lastly, the prefix PA denotes procrustes-alignment, which globally aligns the ground-truth and prediction using procrustes analysis before computing the metric in question

\subsection{Datasets}
We use the following datasets in our experiments.

\noindent\textbf{FreiHAND (FH)} \cite{zimmermann2019freihand} consists of 32'560 frames captured with green screen background in the training set, as well as real backgrounds in the test set. Its final evaluation is performed online, hence we do not have access to the ground-truth for the test set. We use the FH dataset for all supervised and self-supervised training and report the absolute as well as the procrustes-aligned EPE and AUC. 

\noindent\textbf{YouTube3DHands (YT3D)} \cite{youtubehand} consists of in-the-wild images, with automatically acquired  3D annotations via key point detection from OpenPose \cite{openpose} and MANO \cite{romero2017mano} fitting. It contains 47'125 in-the-wild frames.
We use the YT3D dataset exclusively for self-supervised representation learning. YT3D contains only 3D vertices and no camera intrinsic information, hence we report the procrustes-aligned EPE and 2D pixel error via weak perspective projection. 

\begin{figure}
\begin{center}
\includegraphics[width=1\linewidth]{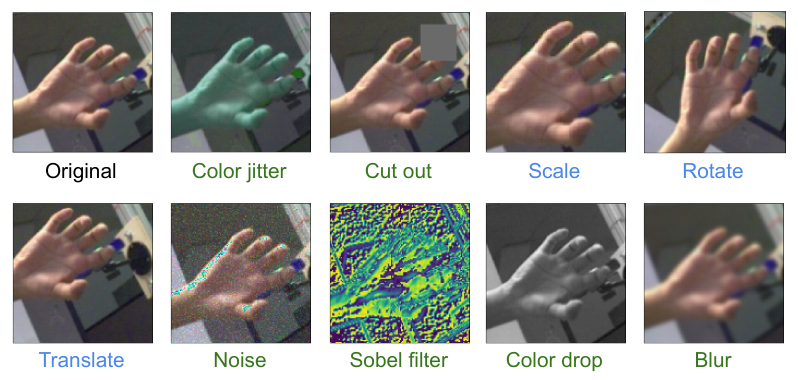}
\end{center}
   \caption{Visualization of transformations evaluated for contrastive learning. Geometric transformations are written in blue whereas appearance transformations are in green. The original sample is taken from FH.}
\label{fig:augmentation}
\end{figure}
\begin{figure}
\begin{subfigure}{\linewidth}
  \centering
  \includegraphics[width=\linewidth]{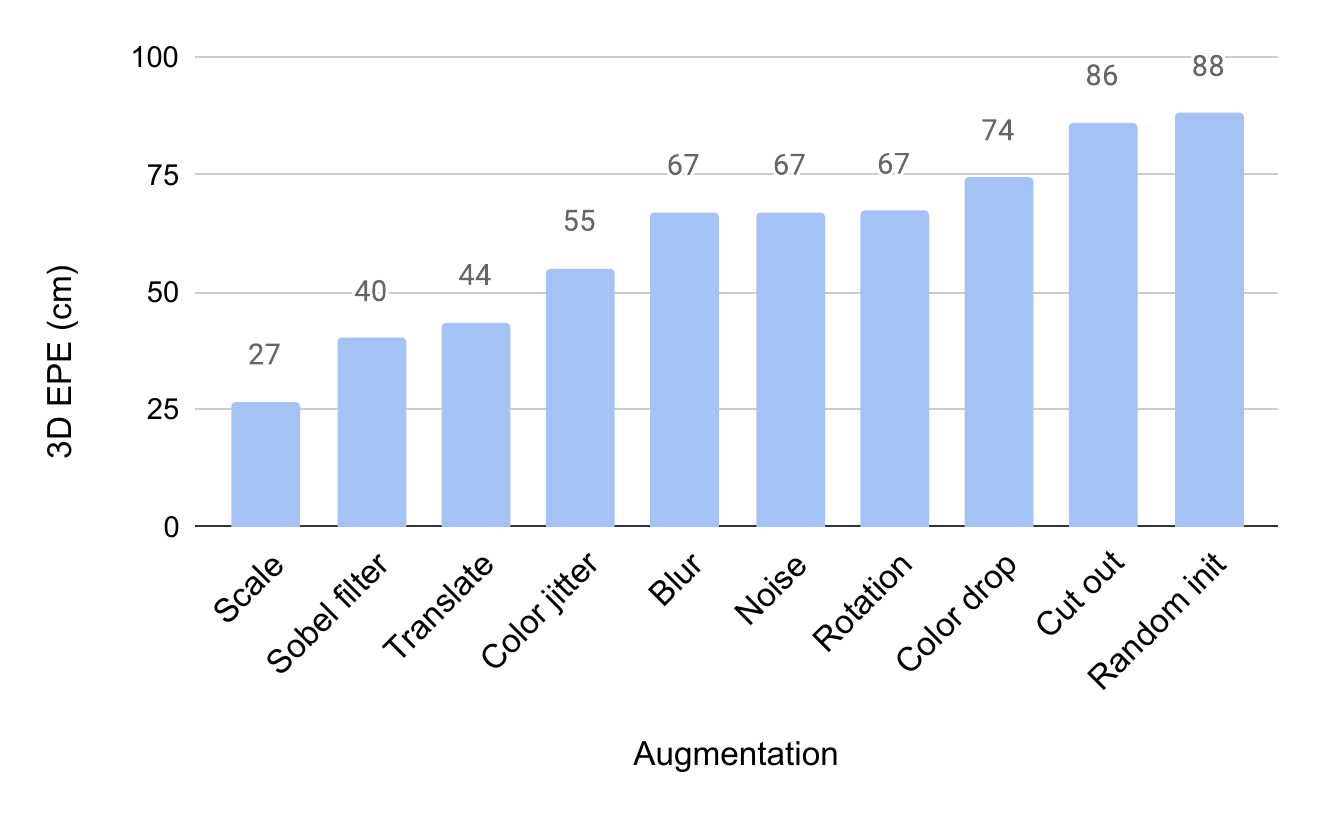}  
  \caption{}
  \label{fig:augmentation_ablation_simclr}
\end{subfigure}
\begin{subfigure}{\linewidth}
  \centering
  \includegraphics[width=\linewidth]{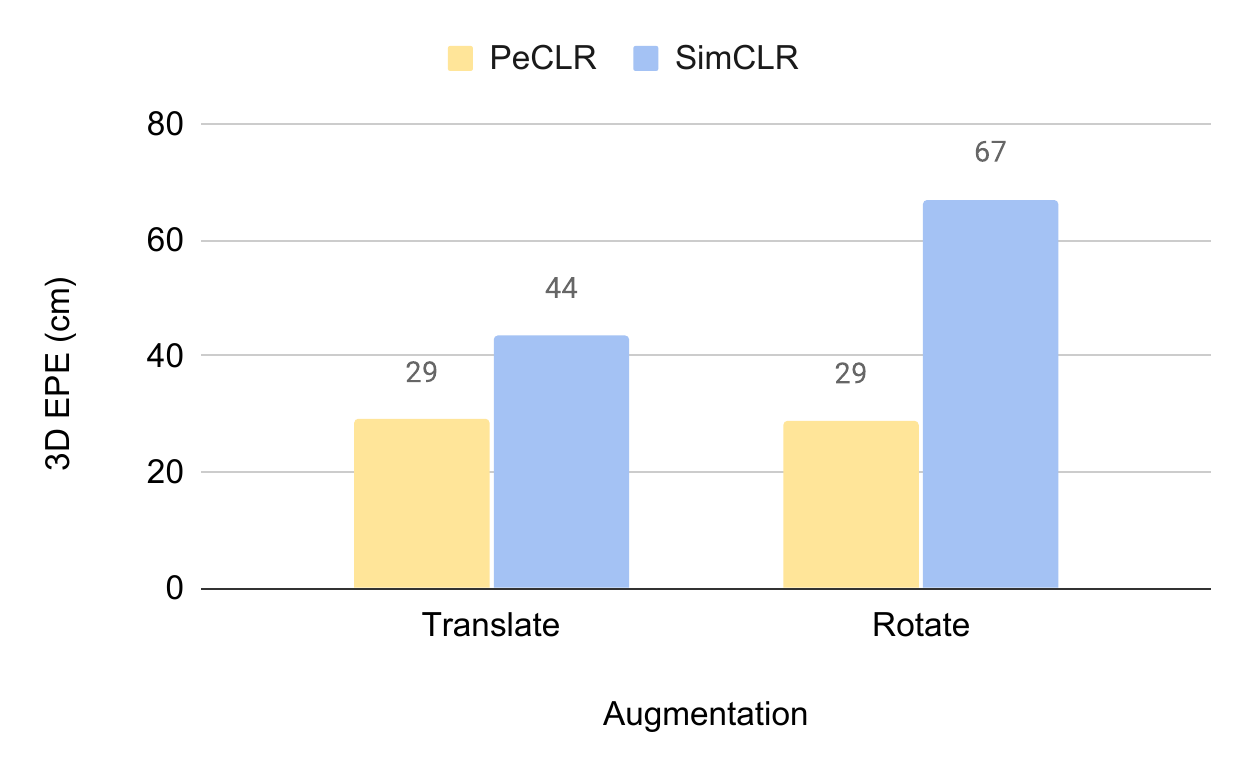}  
  \caption{}
  \label{fig:equivariance_comparison}
\end{subfigure}
\caption{a) The feature representation power of individual augmentation as evaluated by an MLP. b) Comparison of \methodname and SimCLR for translation and rotation, showing a notable improvement of $34\%$ and $56\%$ respectively.}

\label{fig:augmentation_ablation}
\end{figure}
\begin{table}
    \centering
    \begin{tabular}{lccc}
    \hline
         \multirow{2}{*}{Method} & 3D EPE $\downarrow$ & AUC $\uparrow$& 2D EPE $\downarrow$ \\
         & (cm)  &  & (px) \\ 
    \hline
    SimCLR      &  16.62       & 0.72     & 12.05    \\
    PeCLR  (\textbf{ours})           &  \bf 16.05    & \bf 0.74  & \bf 10.51 \\
     \hline
    \end{tabular}
    \caption{Comparison of SimCLR and \methodname on FH. The encoders are pre-trained with either SimCLR or \methodname, and are \emph{frozen} during fine-tuning. Both methods use their optimal set of augmentations, as explained in Sec. \ref{subsec:Importance of augmetations}. 
    }
    \label{tab:augmentation_composition}
\end{table}
\subsection{Evaluation of augmentation strategies}
\label{subsec:Importance of augmetations}
To study which set of data augmentations performs best, we first consider various augmentation operations for the representation learning phase. \figref{fig:augmentation} visualizes the studied transformations in our experiment.
We first evaluate individual transformations and then find their best composition.

We conduct the experiment on FH using our own training and validation split ($90\%$ as training and $10\%$ as validation set) and use a RN50 as the encoder. We train two encoders with different objective functions, one using NT-Xent (Eq.\ref{eq:contrastive}) as proposed in SimCLR, and another one making use of our proposed contrastive formulation (Eq.\ref{eq:contrastive_equiv}). To evaluate the learned feature representation, we freeze the encoder and train a two-layer MLP in a fully-supervised manner on 3D hand labels as described in Sec.~\ref{sec:method_hpe}. %

\myparagraph{Individual augmentation} %
\figref{fig:augmentation_ablation} shows the performance errors when individual augmentation is applied. Here the SimCLR framework is used. 
We observe that encoders trained with transformations perform better than random initialization. However, we see that rotation transformation leads to particularly bad performance. As motivated in Sec.~\ref{sec:peclr}, SimCLR promotes invariance under all transformations, including geometric transformation. We hypothesize that the poor performance stems from this invariance property. To verify this, we compare the performance using the equivariant contrastive loss proposed in \methodname and SimCLR's contrastive formulation under two geometric transformations, namely translation and rotation. We emphasize here again that due to the cosine similarity, the effect of scale is eliminated. \figref{fig:equivariance_comparison} shows that for both translation and rotation, \methodname yields significant improvements of $34\%$ and $57\%$ relative to SimCLR, respectively. This results in scale, translation and rotation having the best feature representation as evaluated by the final MLP's accuracy with \methodname. Note that we only promote equivariance for geometric transformation. Therefore, all other appearance-related transformations yield the same performance for \methodname and SimCLR.

\myparagraph{Composite augmentations} Finally, we compare different compositions of transformations. To narrow down the search space, we pick the top-4 performing augmentations from \figref{fig:augmentation_ablation} as candidates. 
We then conduct an exhaustive search over all combinations of the selected candidates %
and empirically find that scale, rotation, translation and color jitter deliver the best performance for \methodname, whereas SimCLR performs best with scale and color jitter. %

We compare \methodname with SimCLR using their respective optimal composition and report the results in \tabref{tab:augmentation_composition}.
Notice that \methodname yields better feature than SimCLR, gaining the improvements of $3.4\%$ in terms of 3D EPE and $12.8\%$ in terms of 2D EPE.
This demonstrates that PeCLR leads to a more effective representation learning approach for hand pose estimation.  

\begin{figure}
  \centering
  \includegraphics[width=\linewidth]{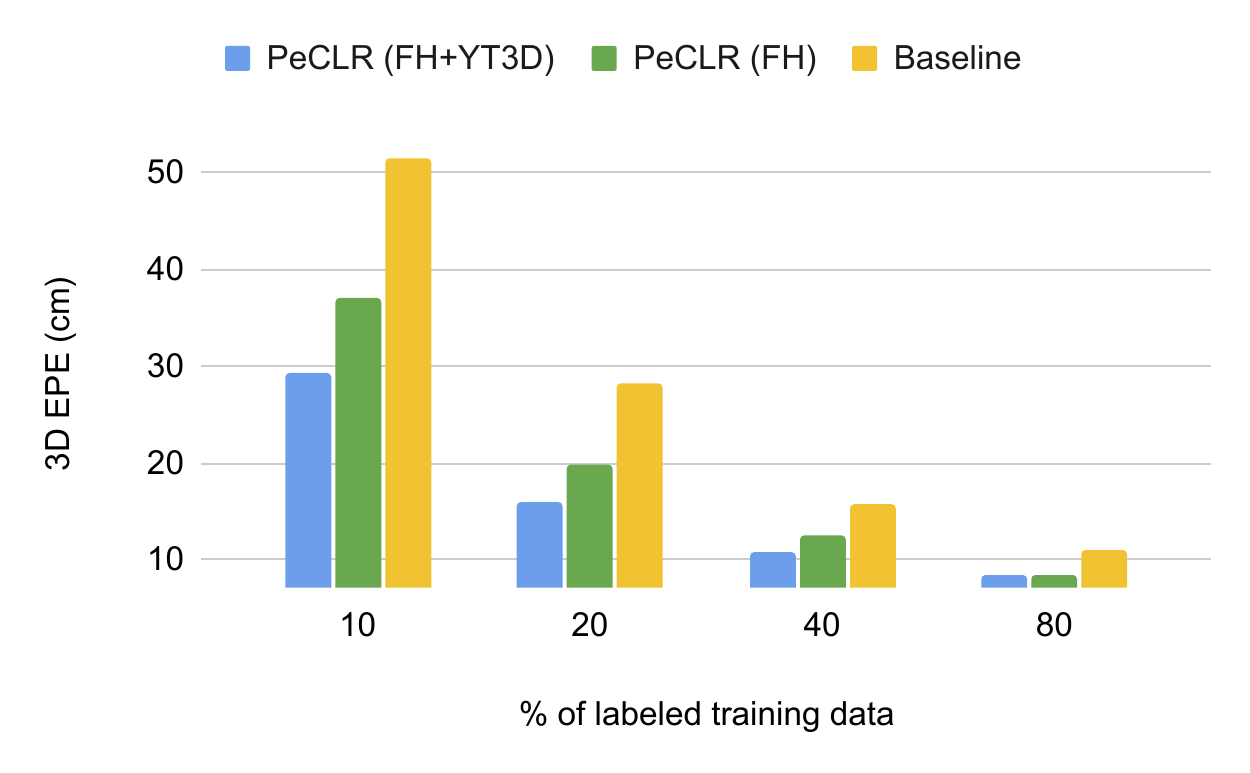}
\caption{Semi-supervised performance on FH. By pre-training with \methodname we achieve greater accuracy in contrast to only training supervised. Adding additional unlabeled data increases this effect.}
\label{fig:semi_sup}
\end{figure}
\subsection{Semi-supervised learning}
\label{subsec:exp_semi_supervised}
In this experiment, we evaluate the efficiency of \methodname in making use of labeled data. To this end, we perform semi-supervised learning on FH with the pre-trained encoder. We use the optimal data augmentation compositions developed in Sec.~\ref{subsec:Importance of augmetations}. As indicated in \cite{chen2020big}, deeper neural networks can make better use of large training data. Therefore, we increase our network capacity and use a RN152 as the encoder in the following. Results and discussion of RN50 can be found in supplementary.

Specifically, we pre-train our encoder on FH with the PeCLR. The encoder is then fine-tuned on varying amounts of labeled data on FH. For clarity, we term the resulting model $\mathbf{M}_{FH}$. To quantify the effectiveness of our proposed pre-training strategy, we compare against a baseline method $\mathbf{M}_b$ that is solely trained on the labeled data of FH, excluding the pre-training step. Finally, to demonstrate the advantage of self-supervised representation learning with large training data, we train a third model, pre-trained on both FH and YT3D, named $\mathbf{M}_{FH+YT3D}$. 

From the results shown in \figref{fig:semi_sup}, we see that $\mathbf{M}_{FH}$, $\mathbf{M}_{FH+YT3D}$ outperform the baseline $\mathbf{M}_{b}$ regardless of the amount of used labels. This result is in accordance with \cite{chen2020big},  confirming that the pre-trained models can increase label efficiency for hand pose estimation. Comparing $\mathbf{M}_{FH+YT3D}$ with $\mathbf{M}_{FH}$, we see that increasing the amount of data during the pre-training phase is beneficial and further decreases the errors. 
These results from $\mathbf{M}_{FH+YT3D}$ and $\mathbf{M}_{b}$ shed light on label-efficiency of the pre-trained strategy. For example, we see that for $20\%$ of labeled data, $\mathbf{M}_{FH+YT3D}$ performs almost on par with $\mathbf{M}_{b}$ using $40\%$ of labeled data %

\subsection{Comparison with state-of-the-art.}
\label{subsec:sota_comparison}
\begin{table}
    \centering
    \begin{tabular}{l|c|c}
   \hline
        Method       & 3D PA-EPE (cm) $\downarrow$ & PA-AUC $\uparrow$  \\
    \hline
            Spurr et al \cite{spurr2020weakly}           & 0.90          &  0.82     \\
            Kulon et al \cite{kulon2020weaklysupervised} & 0.84          &  0.83     \\
            Li et al \cite{li2020exploiting}             & 0.80          & 0.84      \\
            Pose2Mesh \cite{choi2020pose2mesh}           & 0.77          & -         \\
            I2L-MeshNet \cite{moon2020i2lmeshnet}        & 0.74          & -         \\
            \hdashline
            RN50                                        & 0.83          & 0.84      \\
            ~~~ + PeCLR (\bf ours)                      & 0.71          & 0.86      \\
            RN152                                       & 0.74          & 0.85      \\
            ~~~ + PeCLR (\bf ours)                    & \bf 0.66      & \bf 0.87  \\

    \hline 
    
     \hline
    \end{tabular}
    \caption{\textbf{Comparison with SotA}. Standard ResNet models are unable to outperform state-of-the-art methods. By pre-training using \methodname, we yield a performance increase of $14.5\%$ / $10.8\%$ for RN50 and RN152 respectively, resulting in state-of-the-art performance for both networks.}
    \label{tab:SOTAcompare}
\end{table}
With the optimal composition of transformations and representation learning strategy in place, we compare \methodname with current state-of-the-art approaches on the FH dataset. For our method, we use an increased image resolution of $224 \times 224$ pixels and a RN152 as the encoder. The encoder is pre-trained on FH and YT3D with \methodname and fine-tuned supervised on the FH dataset. In addition, we also have a baseline model that is solely trained on FH in a supervised manner. For completeness, we repeat these experiments with a RN50.

\tabref{tab:SOTAcompare} compares our results to the current state-of-the-art. We see that training a RN model supervised only on FH does not outperform the state-of-the-art, even using large model capacity versions such as RN152. We hypothesize that this is due to the comparably small dataset size of FH and thus lack of sufficient labeled data for training. 
However, using \methodname to leverage YT3D in an unsupervised manner improves performance by $14.5\%$ and $10.8\%$ PA-EPE for RN50 and RN152 respectively, outperforming state-of-the-art. Note that all methods in \tabref{tab:SOTAcompare} use highly specialized architectures. In contrast with our formulation, state-of-the-art performance is established in a purely data-driven way. In \figref{fig:qual_results} (bottom) we visualize qualitative results on both our baseline and PeCLR.

\subsection{Cross-dataset analysis}
\label{subsec:exp_cross_dataset}

\begin{table}
    \centering
    \begin{tabular}{lcc}
    \toprule
                                & \bf FH &                                              \\ \midrule

    \multirow{2}{*}{Method}         & 3D EPE $\downarrow$ & \multirow{2}{*}{AUC $\uparrow$}         \\
                                    & (cm)                &                        \\
    \midrule
    RN152                           &   5.05        &   0.34                            \\
    ~~~ + \methodname (\textbf{ours})    &   \bf 4.56    &   \bf 0.36          \\
    \hdashline
    Improvement                     &   9.7 \%        &   5.6 \%                            \\
    \midrule
                                & \bf YT3D &                                             \\ \midrule
    \multirow{2}{*}{Method}         &   3D PA-EPE $\downarrow$ & 2D EPE $\downarrow$      \\
                                    &   (cm)                    &   (px)                \\
    \midrule
    RN152                           &   3.05        &   22.1                           \\
    ~~~ + \methodname (\textbf{ours})    &   \bf 2.88    &   \bf 16.9        \\
    \hdashline
    Improvement             &   5.6 \%       &   23.5 \%                           \\     
    \bottomrule
    \end{tabular}
    \caption{\textbf{Cross-dataset evaluation.} \methodname model with the RN152 architecture is pre-trained on YT3D and FH and then fine-tuned on FH. The model is then evaluated on both FH (top) and YT3D (bottom) test sets. We observe that similar improvements are gained across both datasets.}
    \label{tab:cross_dataset_analysis}
\end{table}
With a large amount of unlabeled training data, we hypothesize that our approach can produce better features that are beneficial for generalization. To verify this, we examine our models of Sec. \ref{subsec:sota_comparison} in a cross-dataset setting. More specifically, we investigate the performance of both models on the YT3D dataset. This sheds light on how the models perform under a domain shift. We emphasize here that neither models are trained supervised on YT3D.

The results in \tabref{tab:cross_dataset_analysis} show that \methodname outperforms the fully-supervised baseline with improvements of $5.6\%$ in 3D EPE and $23.5 \% $ in 2D EPE. These improvements can be observed qualitatively in \figref{fig:qual_results} (top). The results indicate that \methodname provides indeed a promising way forward in using unlabeled data for representation learning and training a model that can be more easily adapted to other data distributions. We note that cross-dataset generalization is seldom reported in the hand pose literature and it is generally assumed to be very challenging for most existing methods while important for real-world applications.

\section{Conclusion}
In this paper we investigate self-supervised contrastive learning for hand pose estimation, making use of %
large unlabeled data for representation learning. We identify a key issue in the standard contrastive loss formulation, where promoting invariance leads to detrimental results for pose estimation. To address this issue, we propose \methodname, a novel method that encourages equivariance for geometric transformations during representation learning. We thoroughly investigate \methodname by comparing the resulting feature representation and demonstrate improved performances of \methodname over SimCLR. We show that our \methodname has high label efficiency by means of semi-supervision. Finally, our \methodname achieves state-of-the-art results on the FreiHAND dataset. %
Lastly, we conduct a cross-dataset analysis on YT3D and show the potential of \methodname %
for cross-domain applications. We believe that \methodname as well as our extensive evaluations can be of benefits to the community, providing a feasible solution to improve generalizability across datasets. We foresee the usage of \methodname on other tasks such as human body pose estimation. 

\myparagraph{Acknowledgments} We are grateful to Thomas Langerak for the aid in figure creation and Marcel Bühler for helpful discussions and comments.

\newpage
{\small
\bibliographystyle{ieee_fullname}
\bibliography{egbib}
}
\newpage

\title{PeCLR: Self-Supervised 3D Hand Pose Estimation from monocular RGB via Equivariant Contrastive Learning\\\large{[Supplementary Material]}}
\author{}
\maketitle

\section{Maximization of sim($\hat{\bm{z}}_i$, $\hat{\bm{z}}_j$)}
In order to minimize the PeCLR cost function:
\begin{equation}
    \loss_{i,j} = - \log \frac{\exp{(\text{sim}((\bm{\tilde{z}}_i, \bm{\tilde{z}}_j)/\tau)}}{\sum_{k=1}^{2N} \mathbbm{1}_{[k\neq i]} \exp{(\text{sim}(\bm{\tilde{z}}_i,\bm{\tilde{z}}_k)/\tau)}},
    \label{eq:contrastive_equiv}
\end{equation}
We need to maximize the numerator $\text{sim}(\tilde{\bm{z}}_i, \tilde{\bm{z}}_j)$ where $\tilde{\bm{z}}_i = (t^g_i)^{-1}\bm{z}_i$. Here we show that this leads to the desired property of equivariance. For convenience, we restate the property of equivariance. Given an image $\bm{I}^n_i$, a transformation $t^g_i$, a model $f$ is equivariant wrt. to $t^g_i$ if:
\begin{equation}
    t^g_i f(\bm{I}^n_i) = f(t^g_i (\bm{I}^n_i)).
    \label{eq:equivariance}
\end{equation}
Recall that for vectors $\bm{x},\bm{y} \in \mathbb{R}^m$, $\max_{\bm{x},\bm{y}}\text{sim}(\bm{x},\bm{y}) = 1$. For a given $\bm{x}$, any $\bm{y} = a\bm{x}, a \in \mathbb{R}$ fulfills this property. Due to this, any scaling effect is removed and $f$ can output any multiple of $\bm{x}$ to satisfy the equation. Hence we assume $t^g_i$ to contain rotation and translation transformations. For simplicity, we set $a = 1$, hence $\bm{x} = \bm{y}$. In the following, we will drop the superscript $g$ and $n$ for ease of notation. Recall that $t_i(\bm{I}) = \bm{I}_i$,  $f(\bm{I}_i) = \bm{z}_i$. We abuse notation slightly, where $t_i$ corresponds to a function performing geometric transformation applied to an image or an affine matrix which can be applied to a vector. In other words, if $t_i$ corresponds to a rotation by $90^\circ$, then $t_i(\bm{I})$ rotates the image by $90^\circ$ and $t_i\bm{x}$ is a matrix vector multiplication, resulting in rotating vector $\bm{x}$ by $90^\circ$. We have:
\begin{equation}
\begin{split}
    \hat{\bm{z}}_i &= \hat{\bm{z}}_j                          \\
    (t_i)^{-1}\bm{z}_i &= (t_j)^{-1}\bm{z}_j    \\
    \bm{z}_i &=  t_i (t_j)^{-1}\bm{z}_j  ~|~ \text{def. } ~ \hat{t}_{ij} := t_i (t_j)^{-1}  \\
    \bm{z}_i &= \hat{t}_{ij}\bm{z}_j                \\
    f(\bm{I}_i) &= \hat{t}_{ij}\bm{z}_j             \\
    f(t_i (t_j)^{-1} \bm{I}_j) &= \hat{t}_{ij}\bm{z}_j \\
    f(\hat{t}_{ij} \bm{I}_j) &= \hat{t}_{ij}f(\bm{I}_j)
\end{split}
\end{equation}
Hence, fulfilling Eq.\ref{eq:contrastive_equiv} leads to the desired property of equivariance in Eq.\ref{eq:equivariance} for rotation and translation in theory. In practice, due to the re-scaling procedure described in Sec. 3 in the main paper (and elaborated on in Sec. \ref{sec:normalizing_translation}), it will be proportionate equivariant to the translation term. Note that this does not take into account clippings that occur when image rotate and translate out of bounds.

\section{Normalizing translation}
\label{sec:normalizing_translation}
\begin{figure}
 \centering
    \includegraphics[width=1.0\columnwidth]{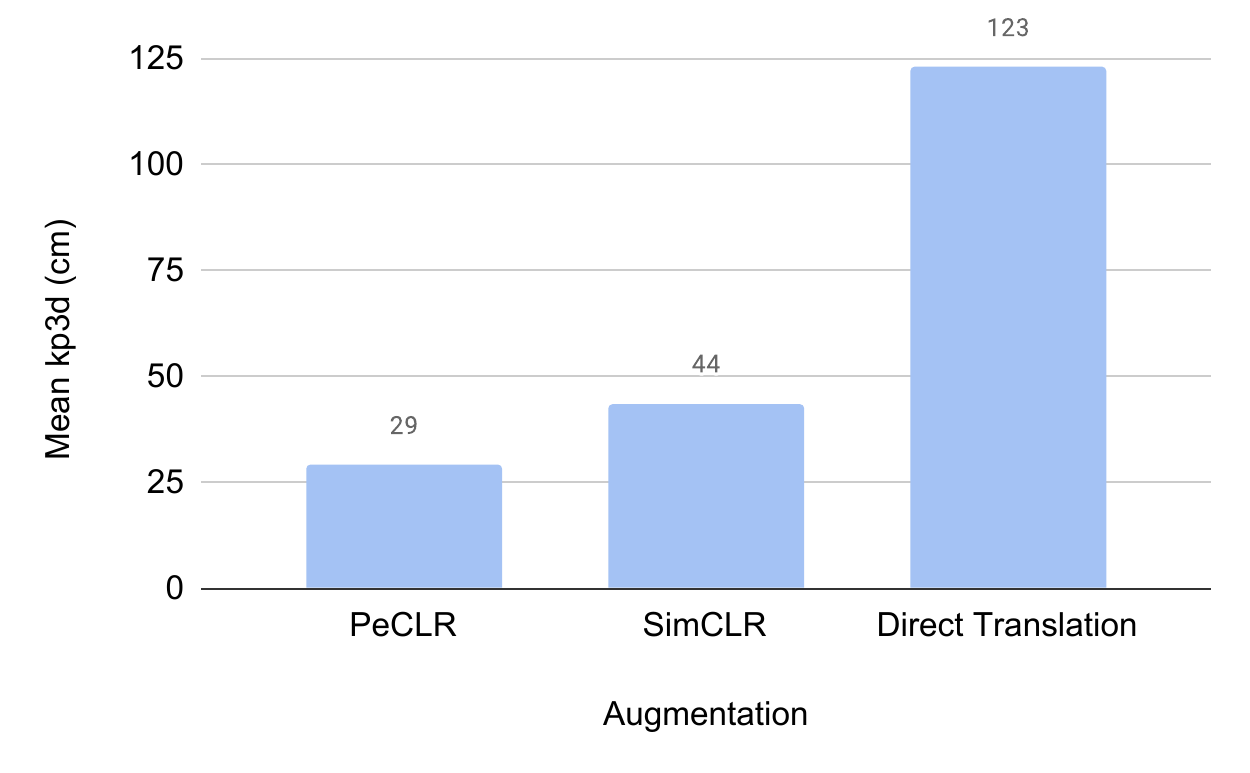}
\caption{Comparing normalized and absolute scale inversion. Here, PeCLR represents the normalized translation and direct translation is a model inverting scale without normalization. For comparison, we show the results on SimCLR too. We see that directly applying a translation has a detrimental effect on performance, performing worse than SimCLR. However, using scaling \methodname leads to superior performance.}
\label{fig:translation_strategy}
\end{figure}
We investigate the effect of our proposed translation normalization procedure. We briefly recap the main motivation behind normalizing translation, as mentioned in Sec. 3 of the main paper. Recall that PeCLR inverts all transformation performed on images in latent space.

Scaling and rotation are transformations that are performed relative to the magnitude. On the other hand, translation is performed in terms of an absolute quantity. Because images are translated in terms of pixels, inverting the translation in latent space by the same quantity may be detrimental. This is due to the differing magnitudes of the pixel and latent space. Therefore we translate the latent space sample $\bm{z}^n$ by a quantity proportional to its magnitude. 

To achieve this, we compute the proportional translation of the image with respect to its size (i.e $\frac{\bm{v}}{L}$, where $\bm{v}$ is the translation vector applied to the image of length $L$). The proportional translation is then multiplied by the magnitude of the latent space, defined as $L_z = \max(\bm{z}_i) - \min(\bm{z}_i)$. In summary, the resulting translation whose inverse is applied to $\bm{z}_i$ is computed as follows:
\begin{equation}
    \bm{\hat{v}} = \frac{\bm{v}}{L} L_z
\end{equation}
Next, we evaluate this choice of normalization. This is done by evaluating the feature representation with and without our proposed normalization in the same manner as in Sec 4.3 in the main paper. \figref{fig:translation_strategy} compares performance of PeCLR with and without translation normalization, as well as SimCLR. We observe that the error of applying direct translation, which omits the normalization scheme results in high errors, performing worse than SimCLR. However, using normalization leads to the best representation, outperforming both SimCLR and direct translation. This quantitatively motivates the use of our normalization procedure.

\section{Training details}
Here, we give more details on the training procedure of PeCLR. 
\textbf{Self-supervised pre-training} is performed for 100 epochs, which is empirically determined to perform best. Following~\cite{SimCLR}, we use ADAM wrapped with LARS and a batch size of 2048. In order to fit the model on a RTX 2080 Ti, we accumulate gradients across smaller batches before back-propagating and use mixed precision for training. Learning rate is set to lr$=\sqrt{\text{batch size}}*1e$-$4$, where a linear warmup is performed for the first 10 epochs. Proceeding that, we use cosine annealing for the remainder of training. While pre-training with multiple datasets, we perform weighted sampling so that a batch consisted of roughly equal amount of samples of each dataset. For PeCLR, we augment the image samples using rotations $r \in[-45,45]$, translation $t \in[-15,15]^2$ and scaling $s \in [0.6,2.0]$. We pick these ranges empirically and find them to perform best. Increasing these ranges degrade performance and sometimes lead to stability issues.
As appearance transformation, we applied color jitter via adjust hue, saturation and brightness. The former two were scaled by a factor $s \in [0.01, 1.0]$ whereas for brightness, we sample scaling factor $s \in [0.5,1.0]$, bias $b \in [5,20]$ and compute $a v_{\text{brightness}} + b$, where $v_{\text{brightness}}$ is the brightness value.

\noindent\textbf{Supervised fine-tuning} is performed for 100 epochs. The adam optimizer with a learning rate of $5e$-$4$ is used in conjunction with cosine annealing. The batch size is set to $128$. Data augmentation is employed, using rotations $r \in [-90,90]$, translation $t \in [-20,20]^2$ and scaling $s \in [0.7,1.3]$.

\section{Semi-supervised learning: RN50}
\begin{figure}
  \centering
  \includegraphics[width=\linewidth]{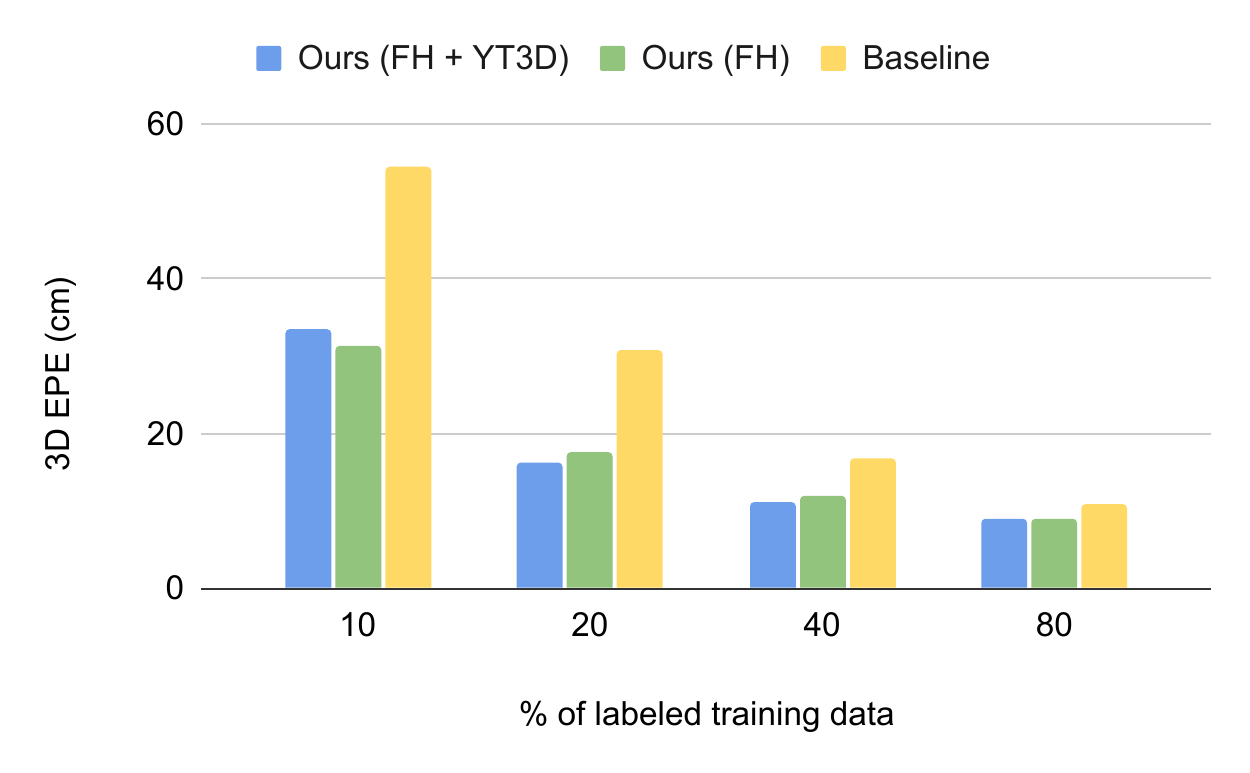}
\caption{Semi-supervised performance on FH using RN50. We observe that by pre-training with \methodname we achieve greater accuracy in contrast to only training supervised. However, the accuracy improvement between using FH and FH+YT3D is smaller as compared to using RN50.}
\label{fig:semi_sup_rn50}
\end{figure}
We conduct the semi-supervised experiment in Sec. 4.4 in the main paper with RN152. In this section, we experiment with RN50 in the same setting as in Sec. 4.3 to show the benefit of using deeper networks.

Following the same experiment steps in Sec. 4.4, we first pre-train our RN50 encoder self-supervised on FH with PeCLR. Next, the encoder is fine-tuned supervised on varying amounts of labeled data from FH. We term the resulting model $M_{FH}$. We compare our proposed pre-training strategy against a baseline method $M_b$, which is trained solely supervised on the labeled data of FH, excluding the pre-training step. Lastly, a third model is trained to demonstrate the advantage of self-supervised representation learning with large training data, pre-trained on both FH and YT3D, named $M_{FH+YT3D}$. 

\figref{fig:semi_sup_rn50} shows the absolute 3D EPE for models across all settings. We observe both PeCLR models $M_{FH},M_{FH+YT3D}$ outperform the baseline model $M_b$ across different labeling percentages. However, the improvement of $M_{FH+YT3D}$ over $M_{FH}$ is lessened when using the RN50 in comparison to the RN152 model. For example, in the $20\%$ labeled setting, by using FH and YT3D for pre-training the RN152 model can gain an improvement of $20\%$ in 3D EPE with respect $M_{FH}$ . On the other hand, the inclusion of additional data lead to an improvement of $8.6\%$ for RN50. This result is consistent with \cite{SimCLR}, which also shows increased performance for larger models.

\section{Further results on other datasets}
\label{sec:sota_other_datasets}
\begin{table*}
    \centering
    \begin{tabular}{lcccc}
    \toprule
                                 \multicolumn{5}{c}{\bf HO-3D}                                  \\ \midrule

     Method                         & 3D PA-EPE (cm)$\downarrow$   & PA-AUC $\uparrow$ & 3D EPE (cm)$\downarrow$ &  AUC $\uparrow$    \\
    \midrule
    Hasson et al. \cite{hasson2019learning}   &   3.18                &   0.46            & 3.27              & 0.44                      \\
    Hampali et al. \cite{honnotate}            &   3.04                &   0.49            & 8.42              & 0.27                      \\ \hdashline
    Supervised                  &   1.33                &   0.74            & 2.85              & 0.50                      \\
    \methodname (\textbf{ours})          &   \bf 1.09            &   \bf 0.78        & \bf 2.26          & \bf 0.58                      \\
    \bottomrule
    \end{tabular}
    \caption{\textbf{HO3D evaluation.} We pre-train a RN50 using PeCLR on YT3D and FH and then fine-tuned supervised on FH. We compare with a model which is solely trained on HO3D (supervised) and note an improvement of $18\%$ in 3D PA-EPE. Performance of \cite{hasson2019learning, honnotate} acquired from \cite{hampali2021website}.}
    \label{tab:ho3d}
\end{table*}
\begin{table}[t]
\centering
\begin{tabular}{lc}
\toprule
\multirow{2}{*}{Dexter+Object}                                      &        \\
                                                                    &  AUC  $\uparrow$ \\\midrule
Mueller (2018)* \cite{mueller2018ganerated}                         & 0.48  \\ 
Spurr (2018) \cite{spurr2018cross}                                   & 0.51  \\
Zimmermann (2018) \cite{zimmerman-3Dpose}                              & 0.57  \\
Baek (2019)* \cite{baek2019pushing}                                 & 0.61  \\
Iqbal (2018)* \cite{iqbal2018hand}                                  & 0.67  \\ 
Boukhayma (2019) \cite{boukhayma20193d}                             & 0.76  \\
Zhang (2019) \cite{zhang2019end}                                    & \bf 0.82  \\
Spurr (2020) \cite{spurr2020weakly}                                 & \bf 0.82  \\ \hdashline
Supervised                                                          & 0.77      \\
PeCLR (Ours)                                                        & 0.81      \\     
\bottomrule
\end{tabular}
\caption{\textbf{Comparison with related work.} We adapt the table from \cite{spurr2020weakly} as it is the most comprehensive comparison with related work. *These works report unaligned results.}
\label{tab:do}
\end{table}
Here we fine-tune models on HO-3D and Dexter+Object. We compare both training solely supervised (our baseline) with using PeCLR pre-training. To provide a better overview, we compare with other related work. We first investigate the results on HO-3D shown in \tabref{tab:ho3d}. The baseline network is solely trained supervised on HO-3D, whereas PeCLR is pre-trained self-supervised on FH and YT3D and fine-tuned supervised on HO-3D. The results are as reported by the online submission system. We report the aligned and unaligned 3D EPE / AUC. On this dataset, we see that our baseline already outperforms related work. PeCLR is capable of pushing the performance even further, yielding an improvement of $18\%$ in aligned EPE. Similar improvements can be found for all other metrics. This demonstrates that PeCLR yields improvement even if the pre-training dataset contains a domain shift with respect to the target dataset.

For Dexter+Object, we use the same network as in Sec. 4.5 in the main paper. \tabref{tab:do} reports the aligned AUC for Dexter+Object dataset. Note that this dataset consists of completely unseen data. We adapt the table from \cite{spurr2020weakly} as it compares a wide range of works. We observe that the baseline network struggles to reach good performance ($0.77$ AUC). However, PeCLR yields improvements of $4.9\%$, almost reaching parity with state-of-the-art ($0.81$ AUC). This experiment indicates that PeCLR results in good cross-domain performance.

\section{Inspecting equivariance of PeCLR and SimCLR}
\begin{figure}
    \centering
    \includegraphics[width=1.0\columnwidth]{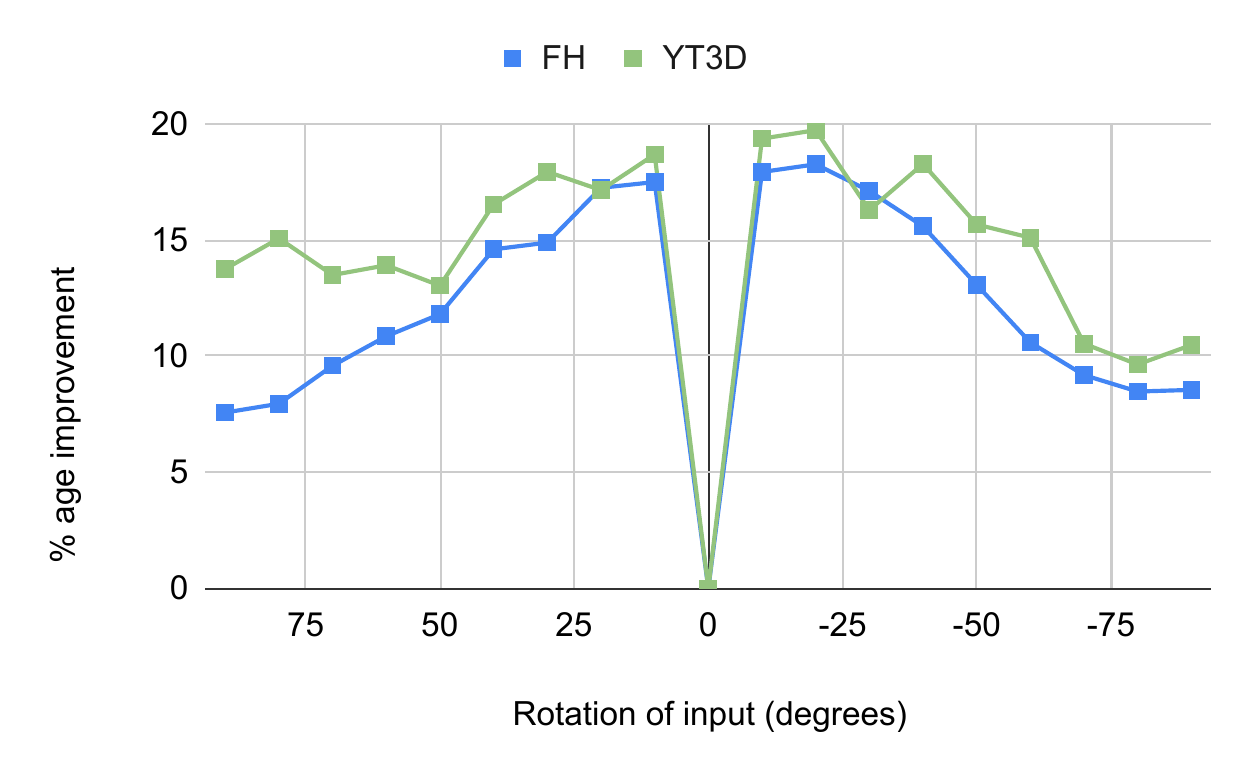}
    \caption{Percentage of improvement for rotational equivariance. Each point denotes the improvement of PeCLR over SimCLR for rotational equivariance, as measured for 2D EPE. We see that across all sampled rotations, PeCLR leads to increased equivariance on both the dataset the model was fine-tuned on (FH) as well as pre-trained (YT3D).}
    \label{fig:equiv_rot_improvement}
\end{figure}

\begin{figure*}
     \centering
     \begin{subfigure}[b]{0.5\textwidth}
        \centering
        \includegraphics[width=1.0\columnwidth]{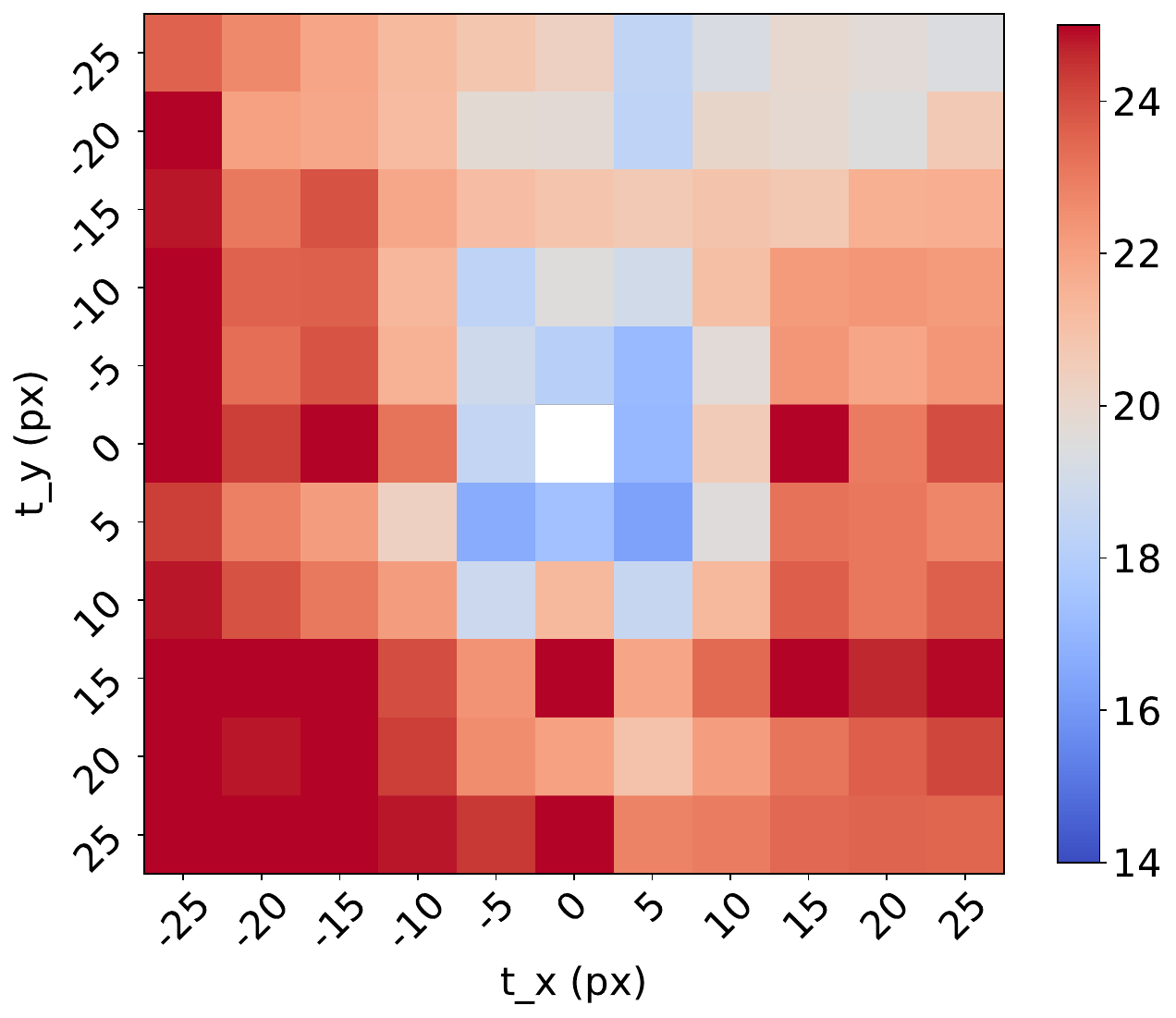}
        \caption{FreiHand}
        \label{fig:quant_equivariance_trans_fh}
     \end{subfigure}%
     \hfill
     \begin{subfigure}[b]{0.5\textwidth}
         \centering
        \includegraphics[width=1.0\columnwidth]{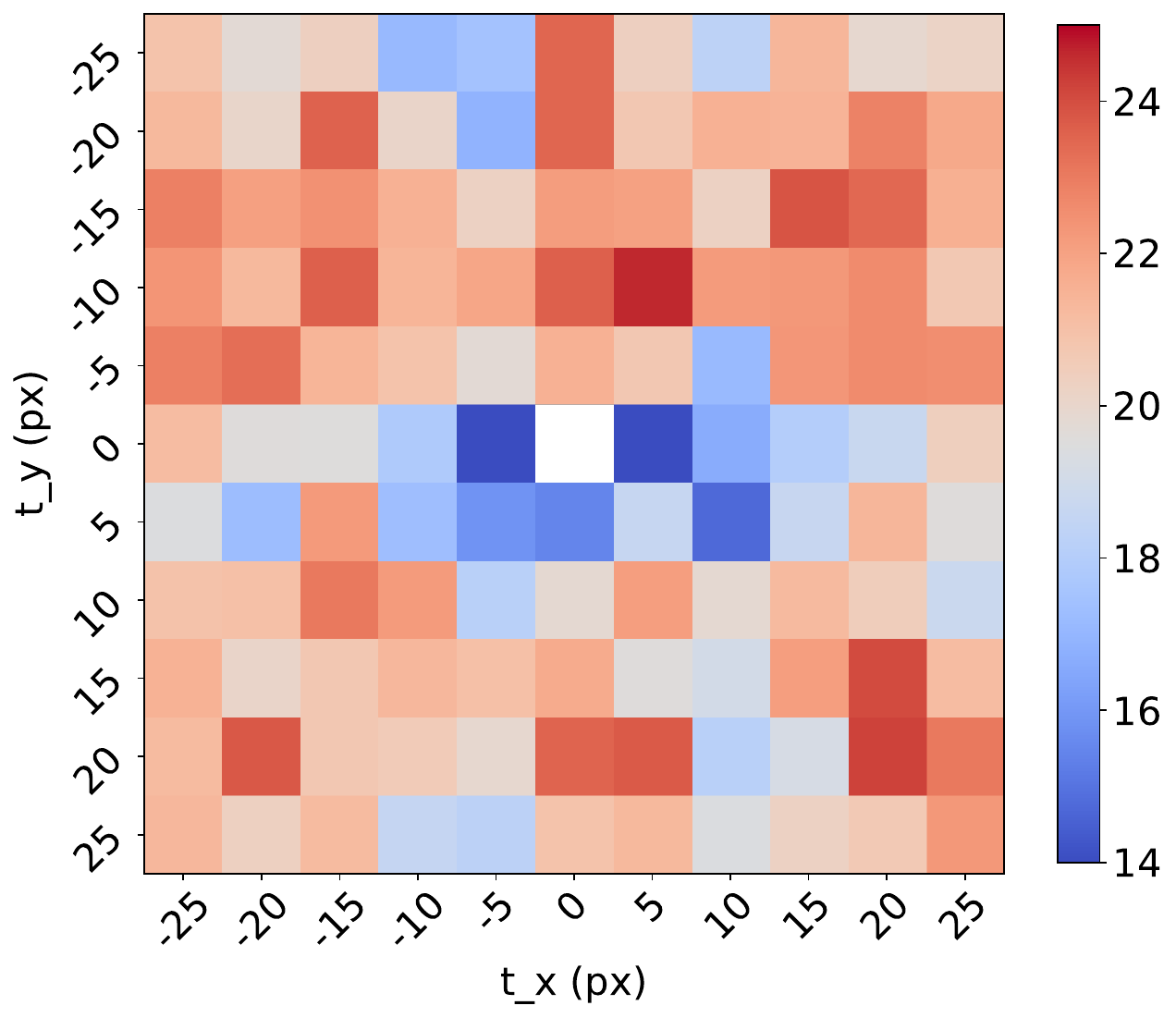}
        \caption{YouTube3DHands}
        \label{fig:quant_equivariance_trans_yt3d}
     \end{subfigure}
\caption{Percentage of improvement for translational equivariance. Each point denotes the improvement of PeCLR over SimCLR for translational equivariance, as measured for 2D EPE. We see that across all sampled translation on the grid, PeCLR leads to increased equivariance on both the dataset the model was fine-tuned on (FH, \figref{fig:quant_equivariance_trans_fh}) as well as pre-trained (YT3D, \figref{fig:quant_equivariance_trans_yt3d}).}
\label{fig:quant_equivariance_trans}
\end{figure*}

\begin{figure}
     \centering
    \includegraphics[width=\columnwidth]{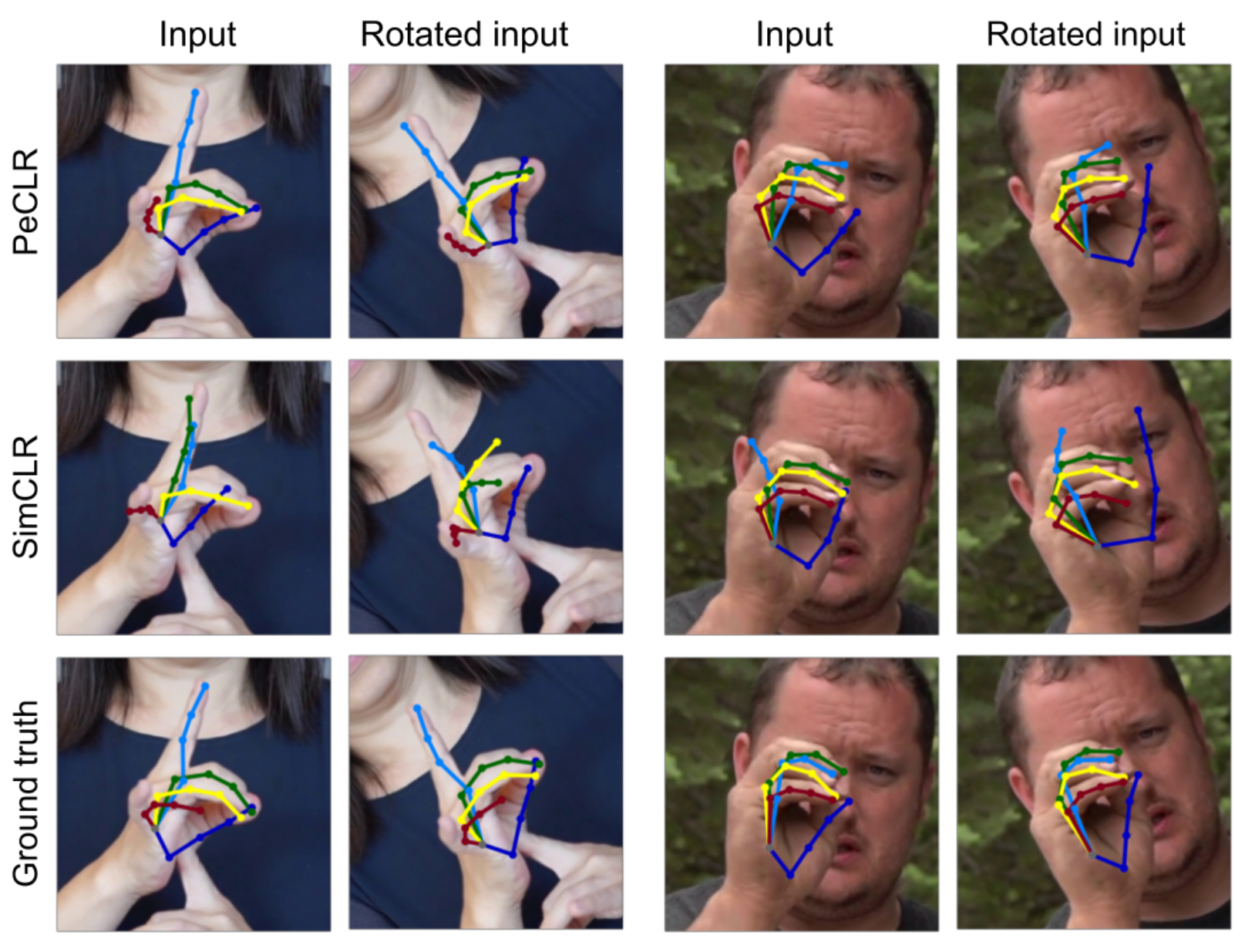}
    \caption{Qualitative samples of SimCLR and PeCLR pre-trained models on YouTube3DHands.}
\label{fig:rebuttal_qual}
\end{figure}

We investigate the equivariance of the resulting model after fine-tuning for both PeCLR and SimCLR. We quantify equivariance by measuring deviations from Eq. \ref{eq:equivariance}. Specifically, we report:
\begin{equation}
    \loss_{equiv}(\bm{I}^n) = ||t^g_i f(\bm{I}^n) - f(t^g_i (\bm{I}^n))||_2.
\end{equation}
We investigate the rotation and translation augmentations since they are affected by PeCLR. To quantify the difference in performance between PeCLR and SimCLR, we visualize the following measure of improvement:
\begin{equation}
  \loss_{improv}(I^n) = \frac{\loss_{equiv}^{SimCLR}(I^n) - \loss_{equiv}^{PeCLR}(I^n)}{\loss_{equiv}^{SimCLR}(I^n)}
\end{equation}
This measure allows quantifying improvement relative to the scale of the error. For a given augmentation, we sample points equidistantly on their respective parameter ranges. For rotation we sample points equidistantly in the range $[-80^{\circ},80^{\circ}]$. For translation, we set the ranges at $[-25,25]^2$. Each point is evaluated on the whole dataset. Here we evaluate on both YT3D and FH. Both models have been pre-trained self-supervised on both datasets and ine-tuned supervised on FH. We first visualize the results for the rotation augmentation as shown in \figref{fig:equiv_rot_improvement}. For both datasets, we see that $\loss_{improv}$ is positive for the entire range tested, indicating that PeCLR performs better on equivariance tasks. The amount of improvement declines as we enter more extreme ranges. The same trend can be observe for both the dataset the models have been fine-tuned on (FH) as well as only pre-trained (YT3D). These results are supported by qualitative analysis, as can be seen in \figref{fig:rebuttal_qual}.

\figref{fig:quant_equivariance_trans} shows the effect of translation on equivariance for both models. Similar to rotation, we observe overall improvement of PeCLR over SimCLR across all ranges sampled, as characterized by $\loss_{improv}$ when more extreme translation is applied. 

This experiment demonstrates that the equivariance property holds even after fine-tuning the network.

\section{Qualitative results}
\begin{figure*}
     \centering
    \includegraphics[width=1.8\columnwidth]{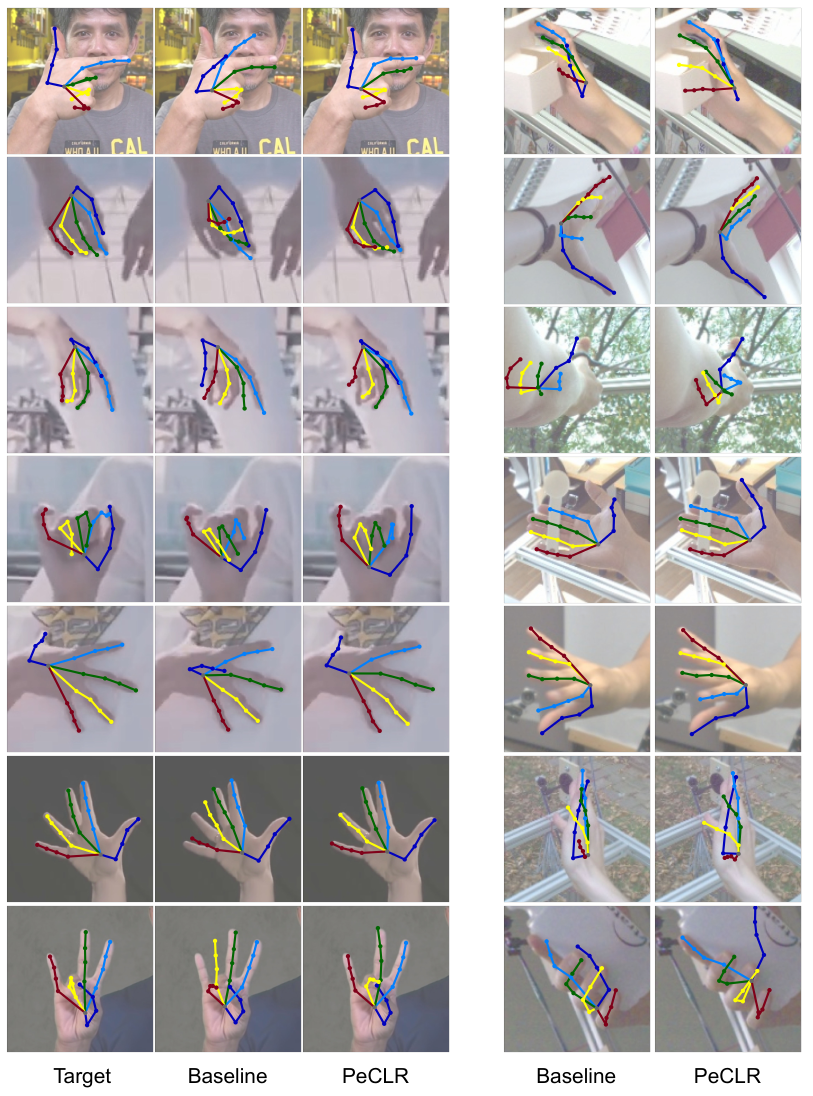}
    \caption{Predictions are shown on the test sets of YT3D (left) and FH (right) without (Baseline) or with PeCLR pre-training. Note that the ground truth of the test set is not publicly available for FH, thus we only visualize the predictions.}
\label{fig:quality analysis}
\end{figure*}

\begin{figure*}
     \centering
    \includegraphics[width=1.8\columnwidth]{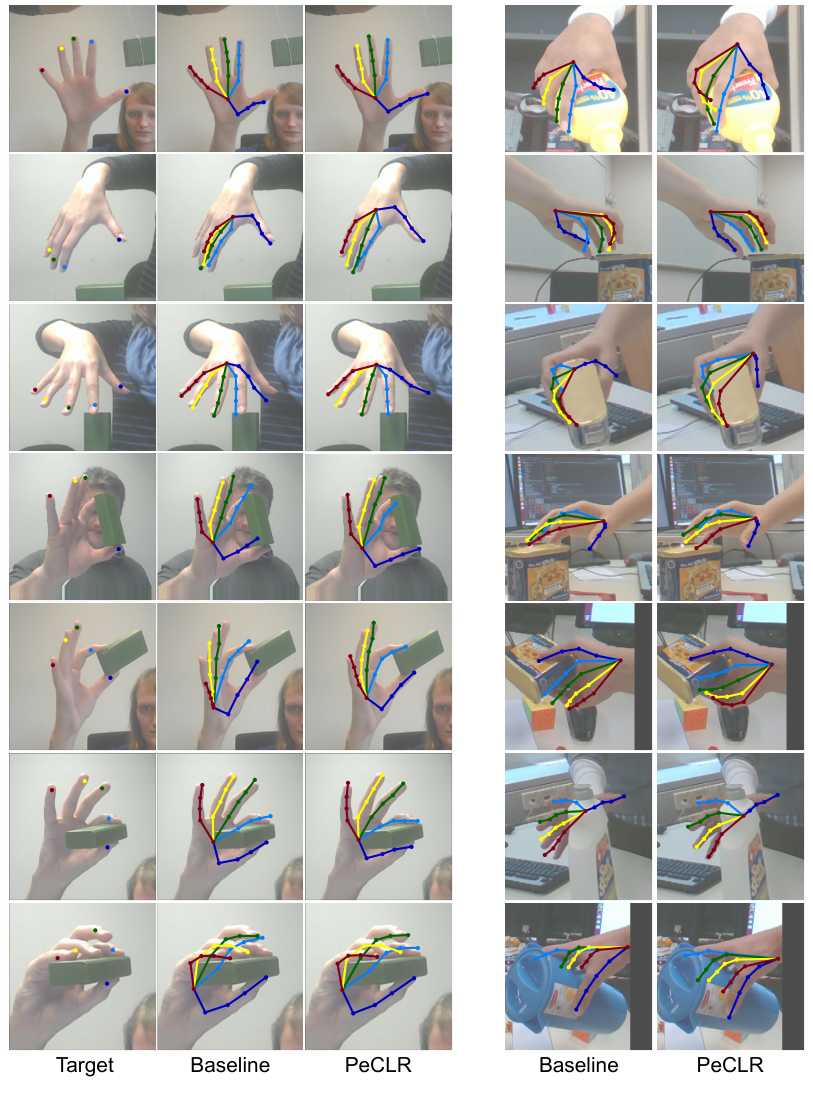}
    \caption{Predictions are shown on the test sets of D+O (left) and HO3D (right) without (Baseline) or with PeCLR pre-training. Note that the ground truth of the test set is not publicly available for HO3D, thus we only visualize the predictions.}
\label{fig:quality analysis_do_ho3}
\end{figure*}

Here we demonstrate further qualitative results on FH and YT3D. Furthermore, \figref{fig:quality analysis} and \figref{fig:quality analysis_do_ho3} visualize predictions on HO-3D and D+O from the models described in Sec. \ref{sec:sota_other_datasets}. 

% List and number all bibliographical references in 9-point Times,
% single-spaced, at the end of your paper. When referenced in the text,
% enclose the citation number in square brackets, for
% example~\cite{Authors14}.  Where appropriate, include the name(s) of
% editors of referenced books.
\end{document}